\documentclass[10pt,journal,compsoc]{IEEEtran}

\ifCLASSOPTIONcompsoc
  \usepackage[nocompress]{cite}
\else
  \usepackage{cite}
\fi

\ifCLASSINFOpdf
 
\else
  
\fi

\usepackage{xcolor}
\usepackage{array}
\usepackage{graphicx}
\usepackage{multirow}
\usepackage[ruled]{algorithm2e} 
\usepackage{amsmath}
\usepackage{subfiles}
\usepackage{import}
\usepackage{hyperref}

\makeatletter
\g@addto@macro\normalsize{%
  \setlength\abovedisplayskip{0.5pt}
  \setlength\belowdisplayskip{0.5pt}
  \setlength\abovedisplayshortskip{0.5pt}
  \setlength\belowdisplayshortskip{0.5pt}
}
\makeatother

\begin{document}

\title{FPDeep: Scalable Acceleration of CNN Training on Deeply-Pipelined FPGA Clusters}


\author{Tong~Geng*, Tianqi~Wang*\thanks{* marked authors have equal contribution and credit}, Ang~Li, Xi~Jin, and Martin~Herbordt
}

\markboth{Transactions on Computers,~Vol.~14, No.~8, August~2020}%
{Shell \MakeLowercase{\textit{et al.}}: Bare Advanced Demo of IEEEtran.cls for IEEE Computer Society Journals}

\IEEEtitleabstractindextext{%
\begin{abstract}
Deep convolutional Neural Networks (CNNs) have revolutionized numerous applications, but the demand for ever more performance remains unabated. Scaling CNN computations to larger clusters is generally done by distributing tasks in batch mode using methods such as distributed synchronous SGD. Among the issues with this approach is that, to make the distributed cluster work with high utilization, the workload distributed to each node must be large; this implies nontrivial growth in the SGD mini-batch size.
In this paper we propose a framework, called FPDeep, which uses a hybrid of model and layer parallelism to configure distributed reconfigurable clusters to train CNNs. This approach has numerous benefits. First, the design does not suffer from performance loss due to batch size growth. Second, work and storage are balanced among nodes through novel workload and weight partitioning schemes. Part of the mechanism is the surprising finding that it is preferable to store excess weights in neighboring devices rather than in local off-chip memory. Third, the entire system is a fine-grained pipeline. This leads to high parallelism and utilization and also minimizes the time that features need to be cached while waiting for back-propagation. As a result, storage demand is reduced to the point where only on-chip memory is used for the convolution layers. And fourth, we find that the simplest topology, a 1D array, is preferred for interconnecting the FPGAs thus enabling widespread applicability.
We evaluate FPDeep with the Alexnet, VGG-16, and VGG-19 benchmarks. Results show that FPDeep has good scalability to a large number of FPGAs, with the limiting factor being the FPGA-to-FPGA bandwidth. But with 250 Gb/s bidirectional bandwidth per FPGA, which is easily supported by current generation FPGAs, FPDeep performance shows linearity up to 100 FPGAs. Energy efficiency is evaluated with respect to GOPs/J. FPDeep provides, on average, 6.4$\times$ higher energy efficiency than comparable GPU servers.
\end{abstract}

}

\maketitle

\IEEEdisplaynontitleabstractindextext

\IEEEpeerreviewmaketitle

\ifCLASSOPTIONcompsoc
\IEEEraisesectionheading{\section{Introduction}\label{sec:introduction}}
\else
\section{Introduction}
\label{sec:introduction}
\fi

\IEEEPARstart{D}{eep} convolutional neural networks (CNNs) have revolutionized applications such as image classification and object recognition \cite{li2019bstc,sun2016data,zhao2017optimizing,venkataramani2017scaledeep,geng2019o3bnn}. But as there remains an open-ended demand for more complex networks and larger datasets, new computing solutions are critical. A challenging problem is that while large training sets are necessary for good generalization, they are also more computationally expensive. Therefore, nearly all of these neural networks are powered by the stochastic gradient descent algorithm (SGD). 

Traditionally, distributed synchronous stochastic gradient descent (D-SGD) has enabled large-scale CNN training by partitioning SGD mini-batches into smaller data batches that can then be processed in parallel and so accelerate CNN training \cite{goyal2017accurate}. A drawback of this method is scalability: to enable continued high utilization as the number of nodes increases, each node must be allocated an ever-larger workload, which means that the mini-batch size must increase.
Larger mini-batches, however, slow training convergence. Thus, while larger clusters provide increased computation capacity, the training time is not proportionally reduced \cite{goyal2017accurate,keskar2016large}. 
In \cite{keskar2016large} the authors demonstrate that increasing batch size increases improper convergence to sharp minimizers, which, in turn, gives rise to poor generalization and thus causes an increasing gap between test and training accuracy. 
Table \ref{tab:mini} shows the performance of small-batch (SB) and large-batch (LB) variants of ADAM on six networks. Comparing LB and SB, we observe that LB does not decrease the accuracy derived from the training set, but does substantially reduce the testing accuracy. 

Certain methods can somewhat reduce this loss of accuracy -- e.g., using dynamic batch sizes and fine-tuning the learning rate -- but they do not solve the problem \cite{goyal2017accurate}. SB limits the parallelism that can be exploited by high-end computing 
clusters, especially when data parallelism is used; SB is thus rarely used in large-scale training.

\begin{table*}[ht] 
\caption{Performance of small-batch (SB) and large-batch (LB). Note that LB does not decrease training accuracy, but reduces the test accuracy \cite{keskar2016large}} 
 \vspace*{-0.1truein}
\centering 
\begin{tabular}{|m{0.6cm}<{\centering}|m{2cm}<{\centering}|m{1.7cm}<{\centering}|m{2.4cm}<{\centering}|m{2.4cm}<{\centering}|m{2.4cm}<{\centering}|m{2.4cm}<{\centering}|  } 
\hline 
 & &  &   \multicolumn{2}{|c|}{Training Accuracy}& \multicolumn{2}{|c|}{Test Accuracy} \\
\hline 
Name&Network Type & Data Set & SB& LB&SB&LB\\ 
\hline 
$F_i$ & FC               & MNIST     & 99.66\% $\pm$ 0.05\% & 99.92\% $\pm$ 0.01\% & 98.03\% $\pm$ 0.07\% & 97.81\% $\pm$ 0.07\% \\  
\hline 
$F_2$ & FC               & TIMIT     & 99.99\% $\pm$ 0.03\% & 98.35\% $\pm$ 2.08\% & 64.02\% $\pm$ 0.20\% & 59.45\% $\pm$ 1.05\% \\
\hline 
$C_1$ & Shallow Conv     & CIFAR-10  & 99.89\% $\pm$ 0.02\% & 99.66\% $\pm$ 0.20\% & 80.04\% $\pm$ 0.12\% & 77.26\% $\pm$ 0.42\% \\
\hline 
$C_2$ & Deep Conv        & CIFAR-10  & 99.99\% $\pm$ 0.04\% & 99.99\% $\pm$ 0.01\% & 89.24\% $\pm$ 0.12\% & 87.26\% $\pm$ 0.07\% \\
\hline 
$C_3$ & Shallow  Conv    & CIFAR-100 & 99.56\% $\pm$ 0.44\% & 99.88\% $\pm$ 0.30\% & 49.58\% $\pm$ 0.39\% & 46.45\% $\pm$ 0.43\% \\
\hline 
$C_4$ & Deep Conv        & CIFAR-100 & 99.10\% $\pm$ 1.23\% & 99.57\% $\pm$ 1.84\% & 63.03\% $\pm$ 0.50\% & 57.81\% $\pm$ 0.17\% \\
\hline 
\end{tabular} 
\label{tab:mini} 
\vspace*{0.05truein}
\end{table*} 

FPGA clusters are a competitive technology for CNN inference \cite{zhang2016energy,sanaullah2018application,lu2017evaluating,chung2018serving,fowers2018configurable}. For CNN training, however, their efficacy is still an open question; one that is addressed in this work. Previous FPGA  clusters for CNN training have generally worked in batch mode ({\it batch} in the computational sense), which uses the distributed synchronous SGD algorithm just described \cite{zhao2016f,guan2017fp,cong2011customizable,hegde2017caffepresso,moss2017high,lian2016framework}. In this approach, called \textit{Data Parallelism} \cite{ben2018demystifying}, each FPGA executes all layers of the CNN. This is done in sequential order, a layer at a time, with a new layer starting only after the previous layer has completed. Data Parallelism has three significant disadvantages. First, optimal FPGA configurations for different CNN layers vary greatly: either the FPGA is suboptimally configured, or the FPGA needs to be reconfigured repeatedly at run-time.  Second, the storage required for weights and intermediate features is generally large enough that off-chip memory must be used. And third, this entire approach suffers from the scalability problem of the distributed synchronous SGD algorithm already described.

Another method, which we call \textit{Layer Parallelism}, is to daisy-chain multiple FPGAs and map the entire CNN onto a single pipeline. Zhang, et al. \cite{zhang2016energy} used Layer Parallelism to accelerate CNNs using FPGA clusters, but only for inference. Their approach, however, still leaves two problems. First, the pipeline is not seamless; a particular layer might stall until the previous layer finishes. All features must, therefore, be cached until the last feature of a layer is obtained. This requires large storage that necessitates the use of off-chip memory. Second, the computational load varies greatly among layers. A naive workload distribution can result in a large number of idle cycles due to inter-layer dependencies. These two problems are present in both inference and training but have a greater impact on the latter. In training, all features of the hidden layers must be cached until their corresponding errors arrive through Back Propagation (BP), thus requiring much larger memory. And due to BP, the number of operations per layer triples.

We propose FPDeep, a novel FPGA-cluster-based training framework for CNNs that solves the problems just described. FPDeep does this by using a hybrid of layer and model parallelism together with a number of new workload/weight balancing strategies. No reconfiguration is needed: each device computes only certain layers or a part of a single layer; each device is optimized independently with respect to its own computation. The cluster is now a single fine-grained pipeline so the batch size can be arbitrarily small. The amount of data that must be saved is drastically reduced eliminating most off-chip memory accesses. Internode communication is simple and pipeline utilization very high. To the best of our knowledge, our work is the first on CNN training with FPGA-based clusters using this method of parallelism and also the first with fine-grained workload/weight balancing.

The underlying theme of this work is to convert batch parallelism to pipeline parallelism, which has obvious benefits. Parallelism is equal to the depth of the pipeline, in this case, many thousands of stages across the cluster. Communication paths can be short so cycle times are as small as the designer can make them. Communication among devices is direct and contention-free with any latency having no effect on throughput. There is also the aforementioned benefit of having all of the latency reduction applied to individual problem instances and so obviating the algorithmic challenges that come with larger batches.

We find this approach to be effective with performance similar to that of GPU clusters of similar size and technology, but with far better power efficiency. The limiting factor is inter-FPGA bandwidth. But, somewhat surprisingly, we find that a 1D topology suffices and that, even using only six transceivers per FPGA (Stratix-V era), FPDeep achieves linear speed-up to 83 FPGAs. Overall, with 250 Gb/s bidirectional bandwidth per FPGA, easily supported by current generation FPGAs, FPDeep's performance shows linearity up to 100 FPGAs. The main contributions are as follows:

1) The possibility of breaking down the well-known scalability wall of CNN training and demonstrating FPGA clusters to be a competitive technology for CNN training;

2) A novel framework for mapping CNN training logic to distributed FPGA clusters that achieves both high efficiency and scalability; that does not suffer from issues related to mini-batch size; and that needs only a simple interconnection network as is available in any multi-FPGA system with efficient inter-FPGA communication and reasonable bandwidth;

3) A fine-grained pipeline design that minimizes the time that features need to remain available while waiting for back-propagation, thus reducing the storage demand to the point where only on-chip memory is required for the convolution layers;

4) Fine-grained partitioning and mapping methodologies, which provide almost perfect workload and weight balancing among FPGAs; this is done by increasing the flexibility of workload and weight allocation, thus leading to improved utilization: multiple FPGAs can cooperatively compute the same layer, while multiple layers can also be mapped to the same device;

5) An RTL code generator that automatically creates RTL implementations based on the mapping scheme generated by FPDeep.

The organization of this paper is as follows. In Section 2, related work is discussed. In Section 3, the methodology of FPDeep is presented and the workload/parameter partition methods are defined. In Section 4, the overall system architecture is given. In Section 5, the implementation of each FPGA node's accelerator is introduced. In Section 6, the experimental results are presented. Discussion and further work are in Section 7.
\section{Background and Related Work}
In this article we use VGG-16, a widely used neural network in image classification, as an example to demonstrate various FPDeep features.

\subsection{Background}

\begin{figure}
\centering
\includegraphics[width=3.5in]{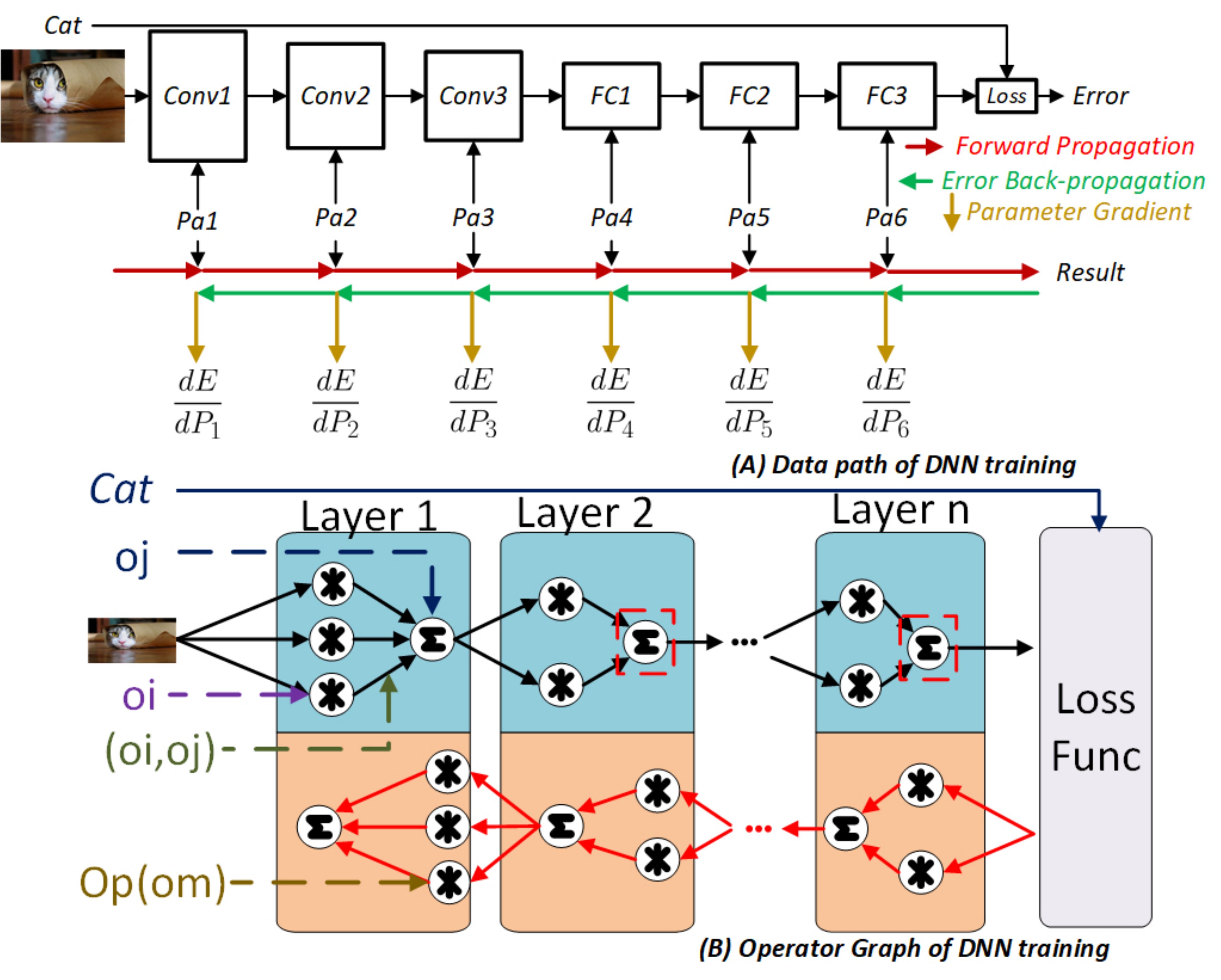}
\vspace*{-0.25truein}
\caption{ (A) Illustration of computations involved in CNN training including datapaths for forward and backward propagation and parameter gradient update. (B) An operator graph is used to represent DNN training. Nodes are operators and edges are tensors.}
\label{DNN_train}
\vspace*{0.05truein}
\end{figure}

The computations for CNN training are shown in Fig. \ref{DNN_train}(A). The red datapath shows Forward-Propagation (FP). It calculates the errors of output features in the final layer. Starting with an input image (Cat), neurons in each layer are evaluated with parameters $Pa_i$. Errors are calculated by comparing inference results to the label in the training dataset. BP has two sub-steps: Error Back-propagation (EB-green) and Parameter Gradient (PG-orange). In EB, errors are back-propagated through the network. In PG errors of each layer are used to calculate gradients of the weights ({$\frac{\partial Err}{\partial Pa_i}$}). 
The convolution kernels are called parameters, the temporal convolution results are called activations. The notation is shown in Table \ref{tab:notation}.

\begin{table}[]
\caption{Definitions of symbols used in the equations}
\vspace*{-0.1truein}
\centering
    \begin{tabular}{|m{1.5cm}<{\centering}|m{6cm}<{\centering}|}
    \hline
    Notation & Description \\ \hline
    A[b][l][c]       & Activation[batch-id][layer-id][channel-id]                                              \\ \hline
    Pa[b][l][o][i]       & Parameter[batch-id][layer-id][output-channel-id][input-channel-id]                      \\ \hline
    dP[b][l][o][i]      & Differential of the parameter[batch-id][layer-id][output-channel-id][input-channel-id]  \\ \hline
    E[b][l][c]       & Error of the layer[batch-id][layer-id][channel-id]                                      \\ \hline
    W                & Size of the activation                          \\ \hline
    K       & Size of the convolution kernel                  \\ \hline
    IC       & Number of the input channels       \\ \hline
    OC       & Number of the output channels                        \\ \hline
    \end{tabular}
\label{tab:notation}
\vspace*{0.05truein}
\end{table}


\begin{figure}
\centering
\includegraphics[width=3.5in]{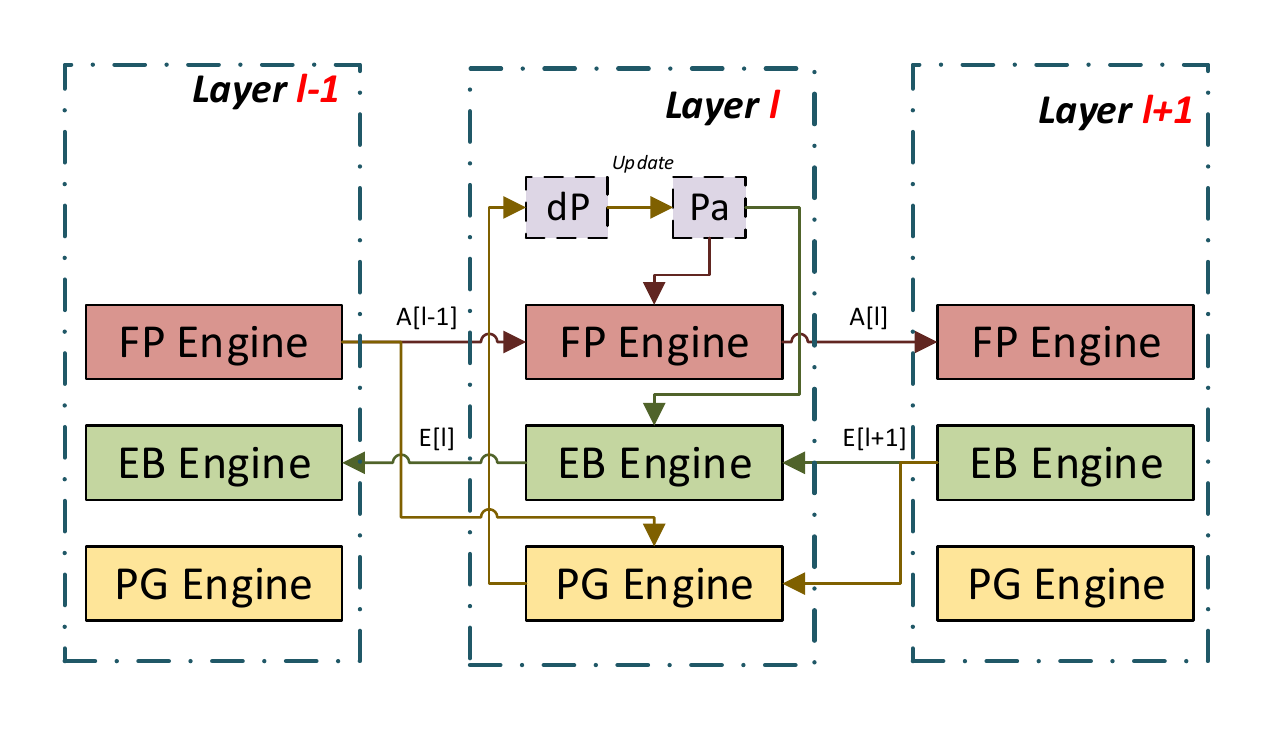}
\vspace*{-0.35truein}
\caption{Execution details of forward and backward propagation with zoom-in on adjacent CONV layers, operations, and data dependencies.}
\label{fig:layer_op}
\vspace*{0.05truein}
\end{figure}

Fig. \ref{fig:layer_op} zooms-in on the FP, EB, and PG operations of two layers. As shown in Tab. \ref{tab:notation}, we use $A$, $Pa$, $dP$, and $E$ to represent Activations, Parameters, Differentials of the Parameters, and errors, respectively. The relationship among of these in CNN training is shown in Eqns. \ref{eqn:act}, \ref{eqn:err}, \ref{eqn:diff} and \ref{eqn:para}.

1. For FP the activation of layer $l$'s channel $c$ is generated by summing all related convolution results of layer $l-1$'s activations and layer $l$'s parameters (Eqn. \ref{eqn:act}).

2. For EB the error of layer $l$'s channel $c$ is generated by summing all related convolution results of layer $l+1$'s errors and layer $l$'s parameters (Eqn. \ref{eqn:err}).

3. For PG the differentials of layer $l$'s parameters are the convolution results of this layer's error and the previous layer's activation (Eqn. \ref{eqn:diff}). The differentials of the parameters are then used ($Pa[b]+dP[b]$) as the next batch's weights ($Pa[b+1]$, Eqn. \ref{eqn:para}).

\begin{equation}
    \label{eqn:act}
    A[b][l][c] = \sum_p A[b][l-1][p] * Pa[b][l][c][p]
\end{equation}

\begin{equation}
    \label{eqn:err}
    E[b][l][c] = \sum_p E[b][l+1][p] * Pa[b][l][p][c]
\end{equation}

\begin{equation}
    \label{eqn:diff}
    dP[b][l][p][q] = A[b][l-1][q] * E[b][l][p]
\end{equation}

\begin{equation}
    \label{eqn:para}
    Pa[b+1][l][p][q] = dP[b][l][p][q] + Pa[b][l][p][q]
\end{equation}


\subsection{Related Work}

Much work has addressed the mapping of inference/training of CNNs to clusters with programmable accelerators, including \cite{blott2017scaling,venkataramani2017scaledeep}. Also, many frameworks and libraries have been deployed, e.g., MXNet \cite{chen2015mxnet}, Caffe \cite{jia2014caffe}, and Tensorflow \cite{abadi2016tensorflow}. These systems hide the complexity of workload decomposition and provide friendly programmer interfaces, including Python, R, and Scala. 
In \cite{mirhoseini2017device}, Google proposed a method that uses reinforcement learning to optimize device placement for the Tensorflow computational graph. \cite{huang2019gpipe} introduced GPipe to solve the problem that the DNN model capacity increases to the point that the model is too big to fit in the memory of a single accelerator. GPipe uses a batch-splitting pipelining algorithm to map AmoebaNet onto eight GPUs. Microsoft proposed PipeDream \cite{huang2019gpipe}, which exploits intra-batch parallelism to train CNNs with GPU clusters. PipeDream uses dynamic programming to find the optimal workload partition. A detailed comparison is given in Sections 3.2 and 4.4.

For FPGA-based clouds, the prior work is more limited. Microsoft's Catapult project \cite{ovtcharov2015accelerating,caulfield2016cloud} includes a parameterized CNN accelerator cluster that can deliver over 1 TFLOPs with very high energy efficiency. Zhang's CDSC FPGA-Enabled Cluster accelerates CNNs on top of Spark and Hadoop \cite{cong2011customizable,zhang2016energy}. In \cite{zhang2016energy}, researchers build a deeply pipelined FPGA cluster with 6 Xilinx VC709 boards to accelerate CNNs. In \cite{zhao2016f}, an FPGA-based framework of CNN training is proposed but focuses mainly on single-FPGA designs.


Most distributed CNN systems, including TensorFlow and CNTK, are based on the distributed synchronous SGD algorithm (Centralized Parallel SGD algorithm - C-PSGD). The Parameter Server Topology \cite{li2014scaling} uses a central parameter node connected with multiple worker nodes. There are multiple bottlenecks including communication load on the central node \cite{lian2017can} and idle time while waiting for straggling worker nodes \cite{chen2016revisiting}. Also, for large-scale clusters, the growth in the SGD mini-batch size limits scalability. Lian, et al. use a decentralized parallel SGD algorithm (D-PSGD) to build a large-scale cluster \cite{lian2017can}. Each node must maintain its own local copy of the model and data duplication is inevitable. 

%
%

Fig. \ref{fig:partition-choice} shows the design space for mapping CNNs onto distributed nodes. We use the terminology introduced by \cite{ben2018demystifying}.
Data parallelism (Fig.\ref{fig:partition-choice}(A)) is the most popular approach in CPU and GPU clouds \cite{chen2015mxnet,abadi2016tensorflow}. It is also widely used in existing FPGA clouds, such as Catapult and CDSC \cite{cong2011customizable}. This method has drawbacks as mentioned in Section I. In CNNs, the configurations of each layer, such as kernel size, pooling size, and stride size, vary greatly, requiring different hardware designs to obtain optimal performance. Thus, FPGAs need to be reconfigured between layers, leading to significant overhead. Also, as each FPGA executes all layers in sequential order, each layer starts only after the previous layer has completed. Thus, for all intermediate features, weights need to be stored to, and loaded from, the host upon completion of a layer, leading to heavy off-chip memory traffic.

Layer Parallelism (Fig. \ref{fig:partition-choice}(B)) maps layers of the CNN onto individual nodes and pipelines the computation. It has been employed by both GPU \cite{huo2018decoupled, huo2018training, wu2016google, jia2018exploring} and FPGA frameworks \cite{zhang2016energy}. In \cite{wu2016google}, each LSTM layer is assigned to a different GPU. 
Since each layer is mapped to a certain GPU, workloads are not balanced among devices. For multi-FPGA systems, Zhang, et al. \cite{zhang2016energy} only addresses inference; also, the parallelism is coarse-grained, the workload is unbalanced, and there is heavy off-chip memory communication. So while Layer Parallelism mitigates some of the problems with batch size and frequent reconfiguration, it suffers from other problems: load balancing and stalls as some nodes wait for others to finish.

In Model Parallelism (Fig. \ref{fig:partition-choice}(C)), weights for each layer are distributed across nodes. Therefore, all intermediate results from all devices must be added up and then broadcast to every device leading to heavy communication. This method has been used for AlexNet \cite{krizhevsky2012imagenet}.

\section{FPDeep Framework}


\subsection{Overview}

An operator graph $G$ (Fig. \ref{DNN_train}(B)) is used to describe the operations in DNN training. Each node $o_i \in G$ is an operator (e.g., matrix multiplication or active function), and each edge $(o_i,o_j) \in G$ is a tensor (an $n$-dimensional array) that is an output of $o_i$ and an input of $o_j$. Each node $o_m$ has a weight $Op(o_m)$. Hardware constraint parameters (Tab. \ref{tab:hard-const}) are used to describe all available hardware devices.


\begin{table}[]
    \caption{Hardware Constraint Parameters}
\vspace*{-0.1truein}
    \centering
        \begin{tabular}{|m{1.6cm}<{\centering}|m{6.4cm}<{\centering}|}
        \hline
        Notation & Description \\ \hline
        Device Num       & Number of FPGA devices in the cluster                       \\ \hline
        $LUT_{max}$      & Number of Look-up table per FPGA device                     \\ \hline
        $FF_{max}$       & Number of Flip-flop per FPGA device                         \\ \hline
        $BRAM_{max}$     & Number of Block-RAM per FPGA device                         \\ \hline
        $DSP_{max}$      & Number of DSP-slice per FPGA device                         \\ \hline
        $Trans_{max}$    & Number of available transceiver per FPGA device             \\ \hline
        \end{tabular}
    \label{tab:hard-const}
\vspace*{0.05truein}
\end{table}

\begin{figure} 
\centering
\includegraphics[width=3.5in]{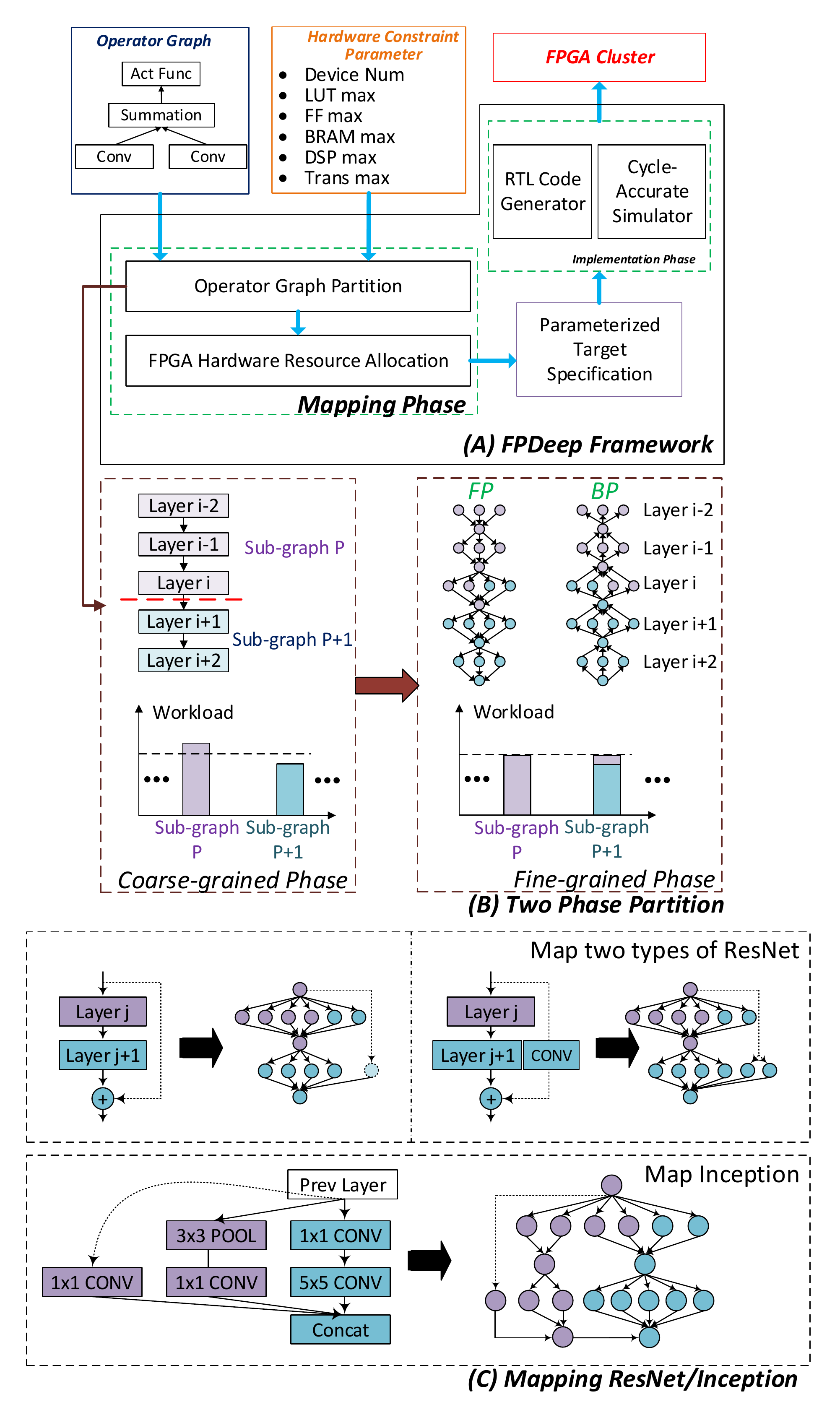}
\vspace*{-0.3truein}
\caption{(A) Overview of the FPDeep Framework. The operator graph and hardware constraints are input parameters. (B) FPDeep contains two phases: mapping and implementation. (C) The proposed DNN operation graph partition methodology with ResNets and Inception.}
\label{fig:framework}
\vspace*{0.05truein}
\end{figure}

FPDeep thus has two sets of input parameters, from the Operator Graph and the Hardware Constraint Parameters. The whole framework contains two parts:  mapping and implementation (Fig.\ref{fig:framework}(A)). The Mapping Framework partitions the operator graph into several fine-grained segments and maps them onto FPGA clusters so that every FPGA gets a balanced workload and parameters.  In the Implementation Framework, the RTL generator creates RTL implementations for each FPGA based on the parameterized mapping, and a cycle-accurate simulator gives measures of throughput, bandwidth demand, and percent idle stages.

\subsection{Operator Graph Partitioning Methodology}

As mentioned in Section 2, DNN training includes FP, EB, and PG phases. The inter-phase data dependencies lead to a complex operator graph $G$. General graph partitioning methods, such as Google's Reinforcement Learning (RL) method \cite{mirhoseini2017device}, are useful approaches in the partitioning tasks of DNN training, but not efficient enough. Finding the global optimal solution of graph partitioning is NP-hard, making the time to find the best partition comparable to the DNN training time \cite{narayanan2019pipedream}. FPDeep takes advantage of the fact that DNN training logic can be modeled as a computational pipeline consisting of groups of consecutive layers; this significantly simplifies the optimization algorithm and makes it possible to return the exact solution in polynomial time.

As shown in Fig. \ref{fig:framework}(B), FPDeep graph partitioning works in two phases: 1) Coarse-grained and 2) Fine-grained.

{\bf 1. Coarse-grained phase}: The whole graph $G$ is abstracted and simplified as a one-way graph $\overline{G}$. Each node in $\overline{G}$ presents the workload of forward and backward propagation of a certain layer. The coarse-grained graph $\overline{G}$ is partitioned into multiple (number of FPGAs) sub-graphs with similar sizes of workloads. This simplifies the partitioning process of $\overline{G}$, but results in a coarse-grained partitioning solution with high variance in the workload size distribution of the sub-graphs.

{\bf 2. Fine-grained phase}: Each sub-graph $\overline{G}_i$ of $\overline{G}$ is replaced with the details of forward and backward propagation. As shown in Fig. \ref{fig:framework}, $\overline{G}_i$ is presented with a finer tiling unit with nodes representing convolution operations at different channels. In this phase, FPDeep performs reallocation of FPGA resources in a finer-grained manner to reduce the variance of workload distribution in phase 1. In Section 6, we showcase the proposed graph partitioning method with practical DNNs e.g. AlexNet, VGG-16/19.

The proposed CNN training graph partitioning method is useful not only for Feed-Forward DNNs (FFDNNs), but also DNNs with more complex topologies, e.g. Residual Neural Networks, Inception, as they can also be modeled as pipelined groups of consecutive layers with some extra abstractions (Fig. \ref{fig:framework}(C)). For example, the parallel convolution and pooling kernels in Inception can be treated as additional output channels of a CONV layer in FFDNNs followed by distributed and pipelined concatenation kernels for data reduction. For Residual Neural Networks, a shortcut can be treated as extra channels of the convolution kernels being bypassed by the shortcut. Support for DNNs with more general topologies will be included in the next-generation FPDeep.

\subsection{Design Choices in Operator Graph Partitioning}


As shown in Fig. \ref{fig:partition-choice}, we can use a cube to represent a node in the operator graph. For each operator, there are three parallelizable dimensions: Sample, Model, and Layer. All available partition choices are shown in Fig. \ref{fig:partition-choice} and Tab. \ref{tab:op-partition}. Four metrics are used to compare the partitioning methods:

\begin{figure}
    \centering
    \includegraphics[width=3.5in]{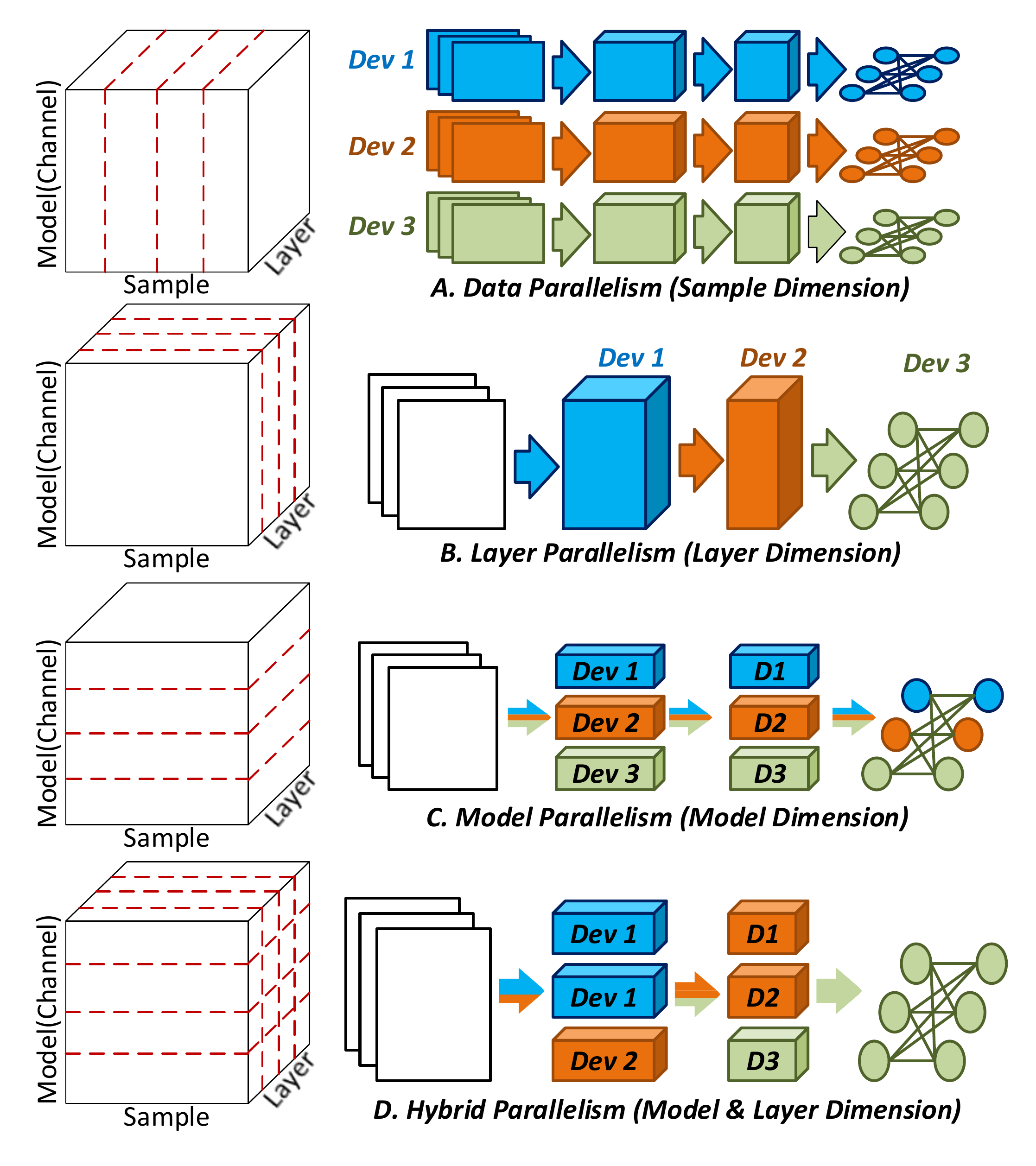}
\vspace*{-0.25truein}
    \caption{Illustration of operator graph partition design choices: (A) Data parallelism, (B) Layer parallelism, (C) Model parallelism, and (D) Hybrid parallelism (Layer + Model)}
    \label{fig:partition-choice}
\vspace*{0.05truein}
\end{figure}

\begin{table}[]
    \caption{Operator Partition Choice. Different operator graph partition design choices make possible different parallelizability methods.}
\vspace{-0.1truein}
    \centering
    \begin{tabular}{|l|l|l|l|}
    \hline
                            & \multicolumn{3}{l|}{Parallelizability Method} \\ \hline
                            & Sample         & Layer         & Model        \\ \hline
    Data Parallelism (DP)   & Y              & N             & N            \\ \hline
    Layer Parallelism (LP)  & N              & Y             & N            \\ \hline
    Model Parallelism (MP)  & N              & N             & Y            \\ \hline
    Hybrid Parallelism (HP) & N              & Y             & Y            \\ \hline
    \end{tabular}
    \label{tab:op-partition}
\vspace*{0.05truein}
\end{table}

{\bf 1. FLOP Utilization.} Maximum FLOPs can be achieved when every DSP slice processes a valid operation every clock cycle. Real performance is less than ideal because of workload imbalance or synchronization overhead. FLOP Utilization ($RealFlops / MaxFlops$) captures this behavior.

{\bf 2. Storage Requirement.} During DNN training, model parameters and temporal activations must be stored. The total Storage Requirement determines whether all necessary data can be stored in on-chip memory. 

{\bf 3. Communication Footprint.}  FPGAs need to synchronize data among co-workers. The Communication Footprint specifies the entire communication data throughput of one mini-batch SGD iteration.

{\bf 4. Communication Bandwidth} is the communication footprint divided by the time of one mini-batch SGD iteration. This metric is used to characterize burstiness.

\begin{figure*} 
    \centering
    \includegraphics[width=7.3in]{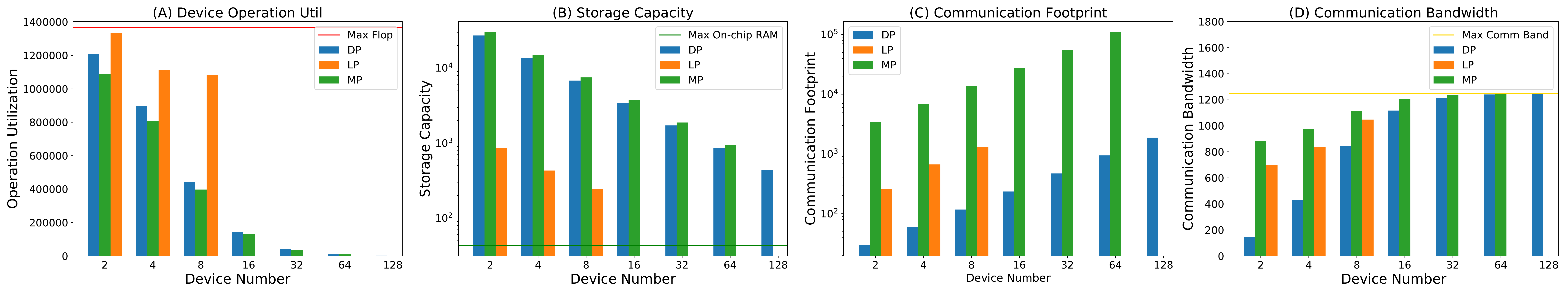}
 \vspace{-0.3truein}
    \caption{Comparison of different operator graph partition methods accounting for four different metrics: FLOP utilization, storage requirement, communication footprint, and average communication bandwidth}
    \label{fig:comp}
\vspace*{0.05truein}
\end{figure*}

Fig. \ref{fig:comp} shows results for these four metrics for different parallelization methods and scales of FPGA clusters. For VGG-16, we set the batch size to 128 so that the DP method works for clusters with fewer than 128 devices. There are 16 layers (13 convolution and 3 fully connection); thus the LP method only works for a cluster with less than 16 devices. Also, the minimum channel count is 64 (Layer CONV-1,2) so the MP method works for clusters with less than 64 devices.

{\bf A. Analysis of Flop Utilization.} LP is the best choice because all devices work in a pipeline manner. However, because of variations in the DNN layer's operation count, LP still suffers from workload imbalance. DP is the second choice. In each mini-batch SGD iteration, DP must synchronize the DNN model globally, which causes serious communication overhead for large scale FPGA clusters. MP is the worst choice since it needs an additional layer for synchronization among different channels.

{\bf B. Analysis of Storage Requirement.} LP is the best choice because of pipelining, which means the cluster does not need to store temporal activations off chip and the DNN model's parameters are distributed. \textbf{Clearly, when the cluster is small, storing all necessary data in the FPGAs' on-chip memory is a challenge}. But when the cluster is large enough, the size of on-chip memory is not a bottleneck. For DP and MP, each FPGA must keep its own copy of the DNN model's parameters. Also, all temporal activations must be maintained in local memory.

{\bf C. Analysis for Communication Footprint.} DP is the best choice because all temporal activations are stored locally and only DNN model parameters need device-to-device communication. LP is the second choice because devices work in a pipeline manner and each device needs to synchronize activations with adjacent devices. MP is worse because it needs to both synchronize parameters globally and synchronize activations among channels.

{\bf D. Analysis for Communication Average Bandwidth.} DP's bandwidth is the lowest due to the centralized burst communication pattern. It only synchronizes the model's parameters after all workers finished their jobs. When the workers are busy, the device-to-device links are idle. In LP, due to all devices working in a pipeline manner, they need to synchronize activations with the adjacent nodes. The communication is stable and the bandwidth of LP is larger than for DP.

{\bf FPDeep Summary:} Rather than DP, LP, or MP, FPDeep uses a hybrid parallel method (Fig. \ref{fig:partition-choice}). It works in a deeply pipelined manner with workload balanced among devices; this improves FLOP Utilization. The balanced allocation policy also reduces the Storage Requirement of the device-to-device activation buffer. As there is no ``free lunch,'' all temporal activations must be transferred among devices. \textbf{Thus the communication bandwidth is the system's bottleneck for large scale clusters}. Tab. \ref{tab:comp} compares the different partition methods. Performance details are presented in Section 6.

\begin{table}[]
    \caption{Qualitative comparison, from 1 (worst) to 4 (best), of different partition methods with respect to various parameters }
\vspace{-0.1truein}
    \centering
    \begin{tabular}{|m{1.2cm}<{\centering}|m{1cm}<{\centering}|m{1cm}<{\centering}|m{1cm}<{\centering}|m{1cm}<{\centering}|m{1cm}<{\centering}|}
    \hline
    Design / Metrics & Parallel Opt & Flop Util & Storage Req & Comm Footprint & Comm Bandwidth \\ \hline
    Data Parallel  & 2            & 2         & 2           & 4              & 4              \\ \hline
    Layer Parallel & 1            & 3         & 3           & 2              & 2              \\ \hline
    Model Parallel & 3            & 1         & 1           & 1              & 1              \\ \hline
    FPDeep         & 4            & 4         & 4           & 3              & 3              \\ \hline
    \end{tabular}
    \label{tab:comp}
\vspace*{0.05truein}
\end{table}

\subsection{Mathematical Model of FPDeep}

As shown in Fig. \ref{fig:framework}, the mapping phase of FPDeep has two parts: operator graph partitioning and FPGA resource allocation. We present a mathematical model for this process. We assume $N$ FPGAs and an operator graph $G$.

\subsubsection{Operator graph partitioning}

In this step, the operator graph $G$ is partitioned into a set of sub-graphs $\mathcal{G} = \{ G_1,G_2 \cdots G_N\}$. Function $Op$ returns the operation count of a sub-graph. For example, $Op(G_i)$ is the operation count of operator graph $G_i$ and $Op_{min}(\mathcal{G})$ is the minimum operation count of the sub-graph set $\mathcal{G}$. Because the FPGA cluster is pipelined, the variance of the sub-graph operation count $V$ should be minimized:

\begin{equation}
\label{eqn:var}
\footnotesize
V = \sum_i \frac{Op(G_i)-Op_{min}(\mathcal{G})}{Op_{min}(\mathcal{G})}
\end{equation}

\subsubsection{FPGA resource allocation}

In this step, the FPGAs' hardware resources are allocated according to the sub-graph set $\mathcal{G}$. The resource allocation step is an optimization problem. The pipeline is constructed from Convolution Engines (CEs), which are used to handle compute-intensive convolution operations, and buffers (Buf), which are used to store CNN model parameters and temporal activations. Convolution engines are composed of 2-D systolic arrays that consume input features from shift-registers. Their design is similar to those in \cite{zhao2016f,wei2017automated}. The FPGAs' resources can be expressed as a tuple: $(LUT, FF, BRAM, DSP)$.

As mentioned above, for large FPGA clusters, the goal is to maximize the cluster's throughput (T), while for small FPGA clusters, the goal is to minimize the storage requirement. In this context, {\it size} is relative and it depends on the ratio of the size of the neural network to the FPGA resources. The constraints lie in the hardware resources at each device and are denoted as $(LUT_{max},FF_{max},BRAM_{max},DSP_{max})$.

{\bf For large clusters} the number of CEs in device $i$ is denoted as $CE^i$. The theoretical maximum performance of these CEs is $Perf(CE^i)$. These convolution engines need buffers $Buf^i$, which is the function of $CE^i$ (Eq. \ref{eqn:buf}). The overall throughput of the cluster is $T$ and depends on the node with the lowest performance (Eq. \ref{eqn:perf}).

\begin{equation}
    \label{eqn:buf}
    \footnotesize
    Buf^i=f_1(CE^i)
\end{equation}

In FPGAs these Convolution Engines or Buffers can be built with hard DSP-slices/Block-RAMs or distributed Lookup-Tables/Flip-Flops. We build some CEs $(\alpha CE^i)$ with hard DSP slices and other CEs $((1-\alpha)CE^i)$ with LUTs/FFs. Similarly, some buffers $(\beta Buf^i)$ are built with hard BRAMs and others $((1-\beta)Buf^i)$ with LUTs/FFs. Equations \ref{eqn:lut}, \ref{eqn:ff}, \ref{eqn:bram}, and \ref{eqn:dsp} define the hardware resource constraints. Functions $f_2,f_3,f_4,f_5,f_6,f_7$ return the consumption of the corresponding hardware resource. The target function for a large cluster is the maximum throughput $T$:

\begin{equation}
    \label{eqn:perf}
    \footnotesize
    T = min(\frac{Op(G_i)}{Perf(CE^i)})
\end{equation}

subject to:

\begin{equation}
\label{eqn:lut}
\footnotesize
LUT^i = f_2(\alpha CE^i,\beta Buf^i,(1-\alpha)CE^i,(1-\beta)Buf^i) \leq LUT_{max},
\end{equation}

\begin{equation}
\label{eqn:ff}
\footnotesize
FF^i = f_3(\alpha CE^i,\beta Buf^i,(1-\alpha)CE^i,(1-\beta)Buf^i) \leq FF_{max}
\end{equation}

\begin{equation}
\label{eqn:bram}
\footnotesize
BRAM^i = f_4(\beta Buf^i) \leq BRAM_{max}
\end{equation}

\begin{equation}
\label{eqn:dsp}
\footnotesize
DSP^i = f_5(\alpha CE^i) \leq DSP_{max}
\end{equation}

{\bf For small clusters} the target function minimizes the storage requirement $S$; the constraints are the same as the large cluster case. To fit all DNN training logic into a small cluster, we propose a method called \textbf{parameter balancing}.

\begin{equation}
    \label{7}
    \footnotesize
    S=max(f_6(\beta Buf^i) + f_7(\beta Buf^i, (1-\beta) Buf^i))
\end{equation}

\begin{figure}[ht!] 
    \centering
    \includegraphics[width=3.5in]{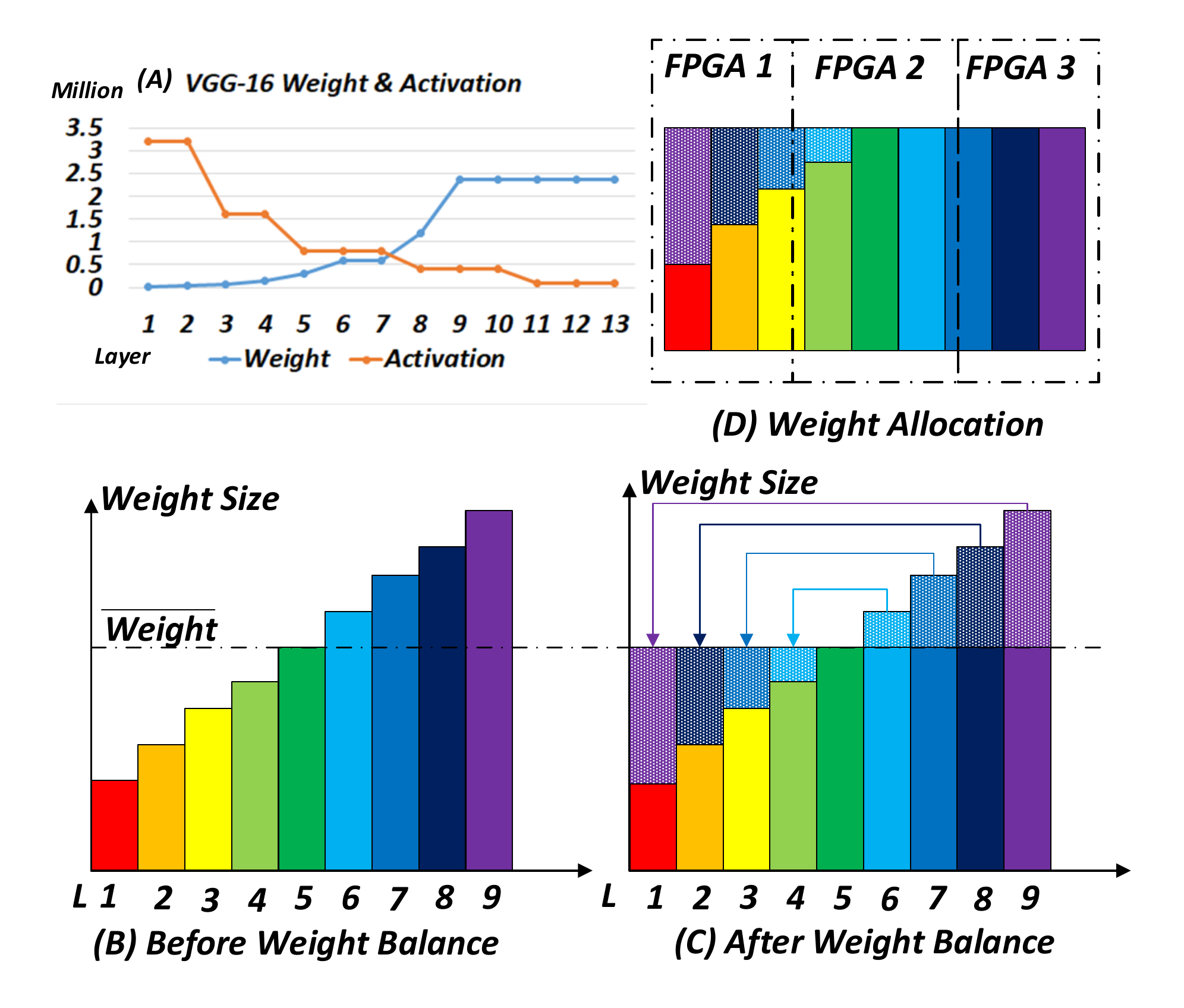}
\vspace{-0.3truein}
    \caption{Parameters and activations for VGG-16}
    \label{WeightBalance}
\vspace*{0.05truein}
\end{figure}

Fig. \ref{WeightBalance}(A) shows the number of model parameters and activations in VGG-16. Observe that from the first to the last layer the number of activations is decreasing while the number of parameters is increasing. The decrease in activations is because the dimensions of the feature maps are reduced by the pooling layers. The increase in the parameters is because the number of input and output channels increases in the later layers. In clusters with small numbers of FPGAs (to accelerate VGG-16) the memory demand of parameters for the later layers increases to the point where the on-chip memories in each FPGA are not big enough to cache the allocated parameters. 

To make enable the mapping of big networks to small clusters of FPGAs, parameter balancing can be used. Figs. \ref{WeightBalance}(B-D) show the method. Simply, parameters from the later layers are stored in FPGAs where there is room, even if those FPGAs are some distance away from where those parameters will eventually be used. For example, the parameters of layer 8 are stored in FPGAs 3 and 1. During computation, the parameters stored in FPGA 1 are transferred to FPGA 3 through the communication network together with activations. Note that the transport of parameters does not tighten the constraint on inter-FPGA communication. This is because the smaller number of activations in the later layers cancels out the added traffic for the parameters. Our experiments demonstrate the benefit of this approach: only on-chip memory is needed for the CONV layers.

\section{System Design}

\subsection{Input/output channel partition implementation}

\begin{figure} 
\centering
\includegraphics[width=3.5in]{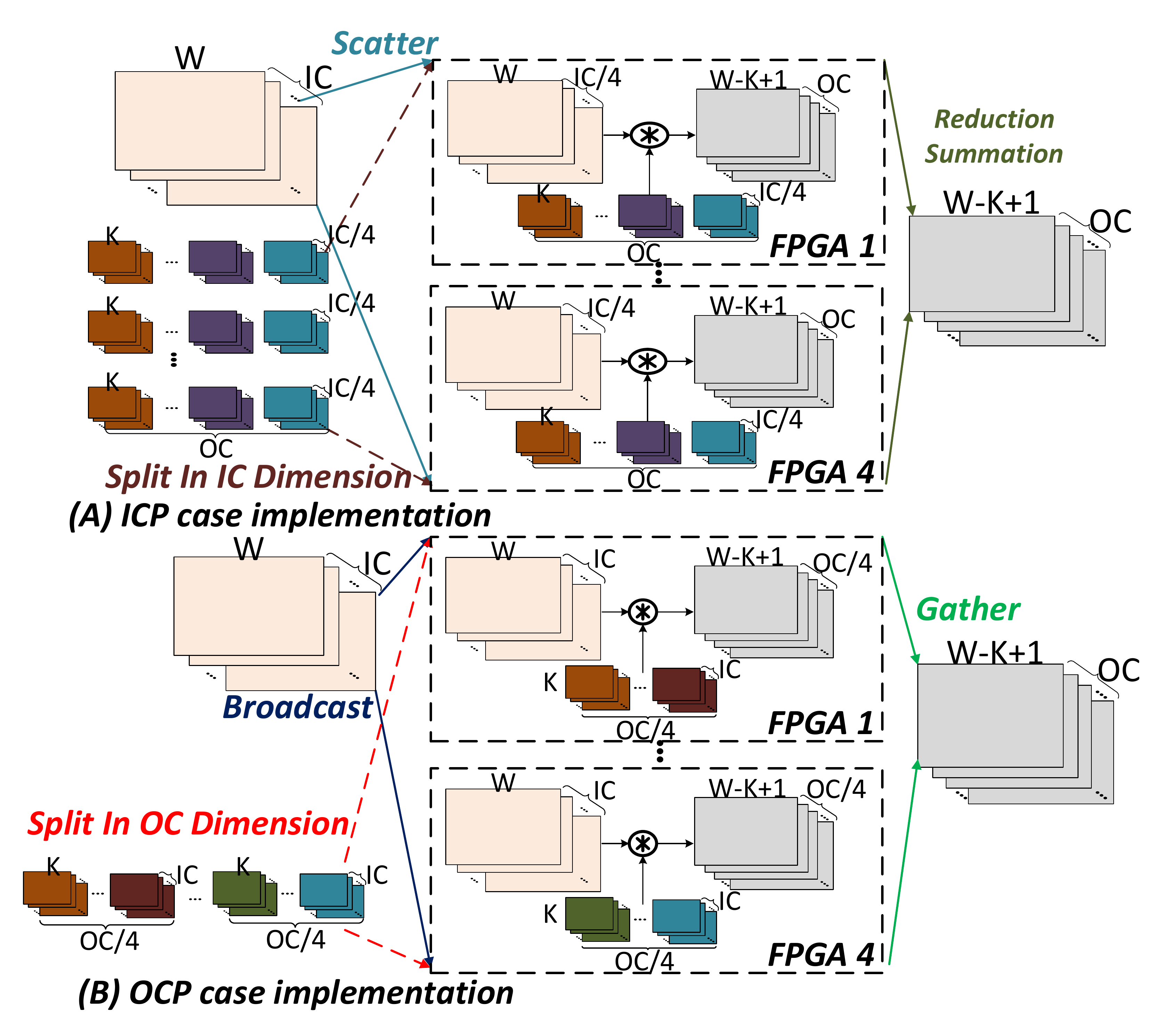}
\vspace*{-0.3truein}
\caption{Partitioning image input/output channels ic/oc}
\label{fig:p_ic_oc}
\vspace*{0.05truein}
\end{figure}

FPDeep uses hybrid parallelism to partition the operator graph. We begin by noting that partitioning in the layer dimension is straightforward (Fig. \ref{fig:partition-choice}(D)). The model dimension is more involved and is done via input/output channel partitioning. As shown in Fig. \ref{fig:p_ic_oc}, each device executes the operations of a fraction of the input/output channels. Input feature maps, along with model parameters, are partitioned in the ic/oc dimension and allocated among FPGA devices. Each device generates the partial results and their sum is the final output activation. 

%

There are two ways to partition the graph, by input (ICP) or by output (OCP) channels. These methods are shown in Figs. \ref{fig:p_ic_oc}(A) and (B), respectively.

1) ICP: Layer 1 is partitioned and mapped to 4 devices. $IC$ input activation channels and corresponding weights are partitioned into 4 segments each containing $IC/4$ channels. Each FPGA receives one of the 4 segments and calculates partial results of activations for all output channels. Each complete output activation is calculated by summing up the related partial results from the 4 FPGAs. 
    
2) OCP: $OC$ output activation channels are partitioned into 4 segments each containing $OC/4$ channels. Each FPGA is responsible for calculating a certain segment of output activations. The 4 segments' results are then gathered. In CNN training, all activations need to remain available while waiting for back-propagation. Therefore, all $IC$ input feature maps are cached in every FPGA. This duplication leads to additional on-chip memory overhead. This defect of OCP does not exist in ICP, so FPDeep prefers to use ICP: OCP is only used when the number of the input channel is too small to provide sufficient parallelism. For example, the first layer of AlexNet only has 3 channels of input features but 96 channels of output features so OCP is used.

\subsection{Dataflow Analysis and Interconnection Topology}

\begin{figure*}
\centering
\includegraphics[width=7.2in]{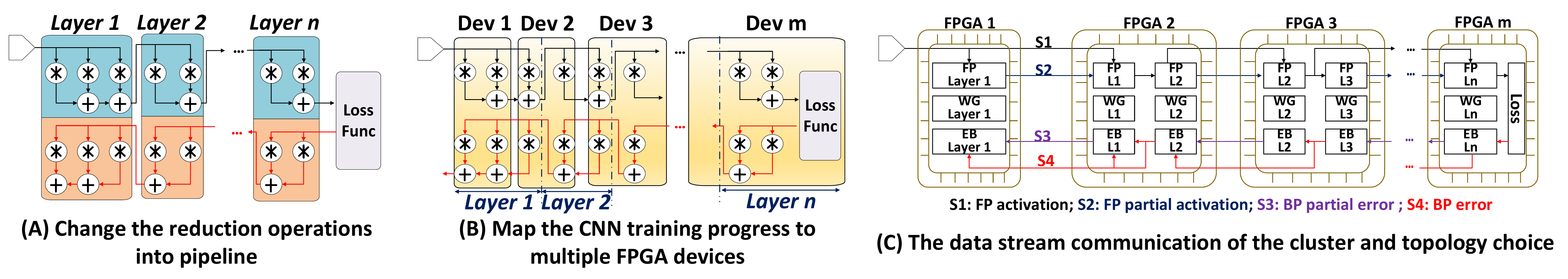}
\vspace{-0.3truein}
\caption{Data flow analysis of CNN training. FPDeep pipelines the reduction operations and maps them to multiple FPGAs.}
\label{fig:partition}
\vspace*{0.05truein}
\end{figure*}

Fig. \ref{fig:partition}(A) shows an $N$ layer CNN mapped to an FPGA cluster with $M$ devices. Each CNN layer contains $O_i$ operations ($i \in [1,N]$). The computation capacity of each device is $C$ operation per second. To balance compute workloads among devices, the workloads are mapped to FPGAs in proportion to the device's compute capacity. Each device needs to execute $\overline{W}=\frac{\sum C_i}{M}$ operations. 

Fig. \ref{fig:partition}(B) zooms in on the CNN training procedure that turns the dataflow into an operator graph as 
the summation of all activation channels are cut into several pieces. Each piece only needs to add the local convolution result $R_i$ to the previous node's intermediate result $R_{i-1}$. The workloads, i.e., the arithmetic operations, of a whole network are partitioned and mapped onto $M$ FPGA nodes in proportion to their computation capacities.
Fig. \ref{fig:partition}(C) shows the data streams in the FPGA cluster.
Note that a 1-D interconnect topology is sufficient. 

\begin{figure}
    \centering
    \includegraphics[width=3.5in]{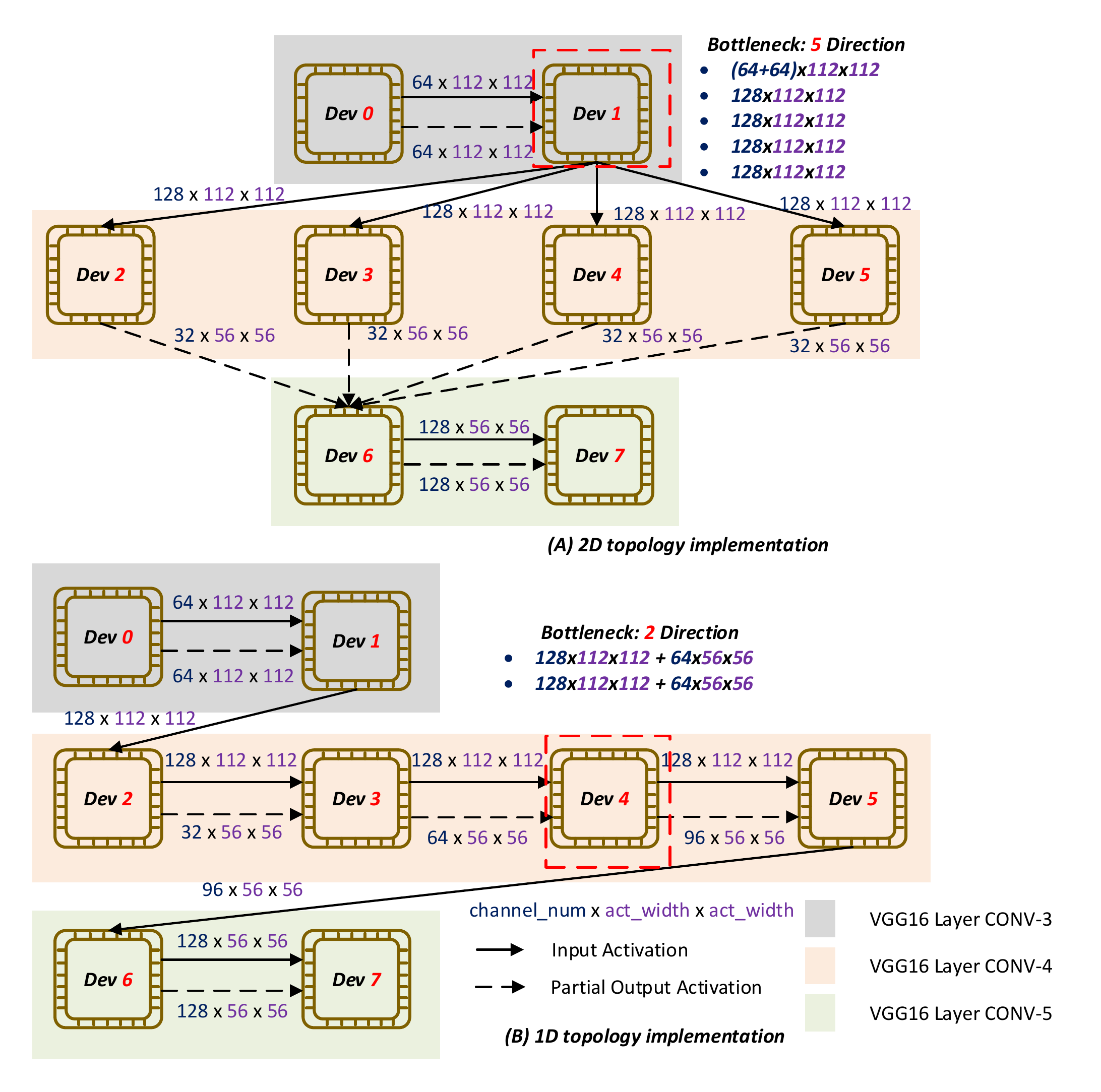}
\vspace{-0.3truein}
    \caption{1D-2D topology design choice: while 2D seems the obvious choice, clearly 1D has better performance}
    \label{fig:top_choice}
\vspace*{0.05truein}
\end{figure}

Fig. \ref{fig:top_choice} illustrates the topology design choice by mapping VGG-16's CONV-3 - CONV-5 layers onto a cluster with eight VC709 FPGA boards (see Section 6) according to FPDeep's operator graph partition method and FPGA resource allocation policy. The red dotted box marks the communication bottleneck. Let us assume 10 board-to-board interconnections ports. First assume a 2D topology. We see that Dev-1 is the bottleneck: because the degree of Dev-1 is 5, each communication link of Dev-1 only has two ports. With a 1D topology, however, Dev-4 is the bottleneck. Some of the partial output activations need to be duplicated (dotted arrows in and out of Dev-4). But the degree of Dev-4 is only 2 while each communication link has 5 interconnection ports, making the communication more efficient.

\begin{figure}
    \centering
    \includegraphics[width=3.5in]{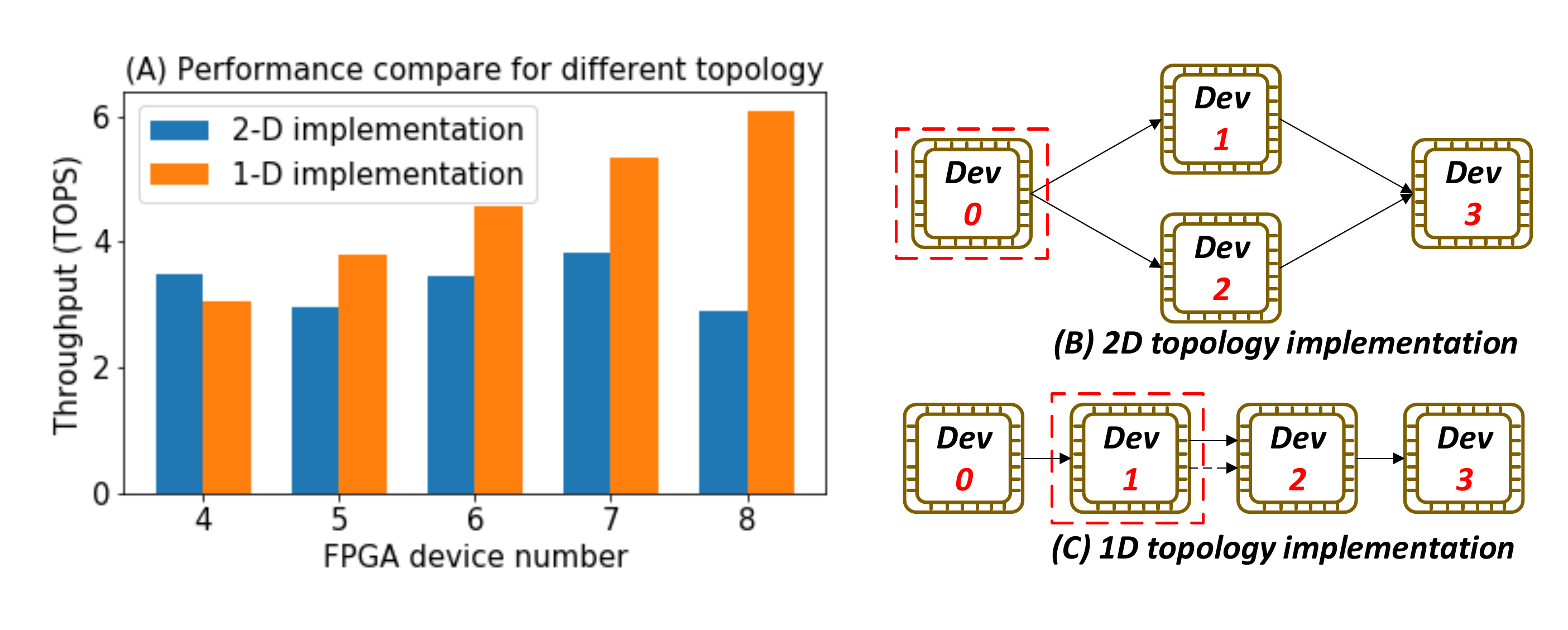}
 \vspace{-0.3truein}
    \caption{1D-2D topology performance comparison}
    \label{fig:top_eval}
\vspace*{0.05truein}
\end{figure}

Fig. \ref{fig:top_eval} shows quantitatively how the choice of topology affects performance. For clusters larger than 5 nodes, the 1D topology is better. As the number of nodes increases, this advantage becomes even more apparent. For clusters with 4 nodes, the 2D topology is better because the degree of the bottleneck device is only 2.

A further advantage of 1D topology versus 2D is its simplicity. With only single links, different dataflow types are multiplexed and easily scheduled. Also, for 2D the reduction operation of each DNN layer is centralized, which incurs significant synchronization overhead and requires more data movement.

Another consideration is that, practically, FPGA accelerator boards almost always have less communication capability than the FPGAs themselves, both in BW and number of ports. This makes the choice of 1D even more crucial. An interesting exception is for accelerator boards with multiple tightly coupled FPGAs. For single boards with, say, four FPGAs, we have already noted that 2D is preferred. For clusters with multiple multi-FPGA boards, because internode connectivity is more limited than intranode, the preferred inter-node topology is again 1D. Within the node, however, the additional {\it in situ} connections remain useful leading to a hierarchical topology.



\subsection{Deep Fine-Grained Pipeline}

\begin{figure} 
\centering
\includegraphics[width=3.6in]{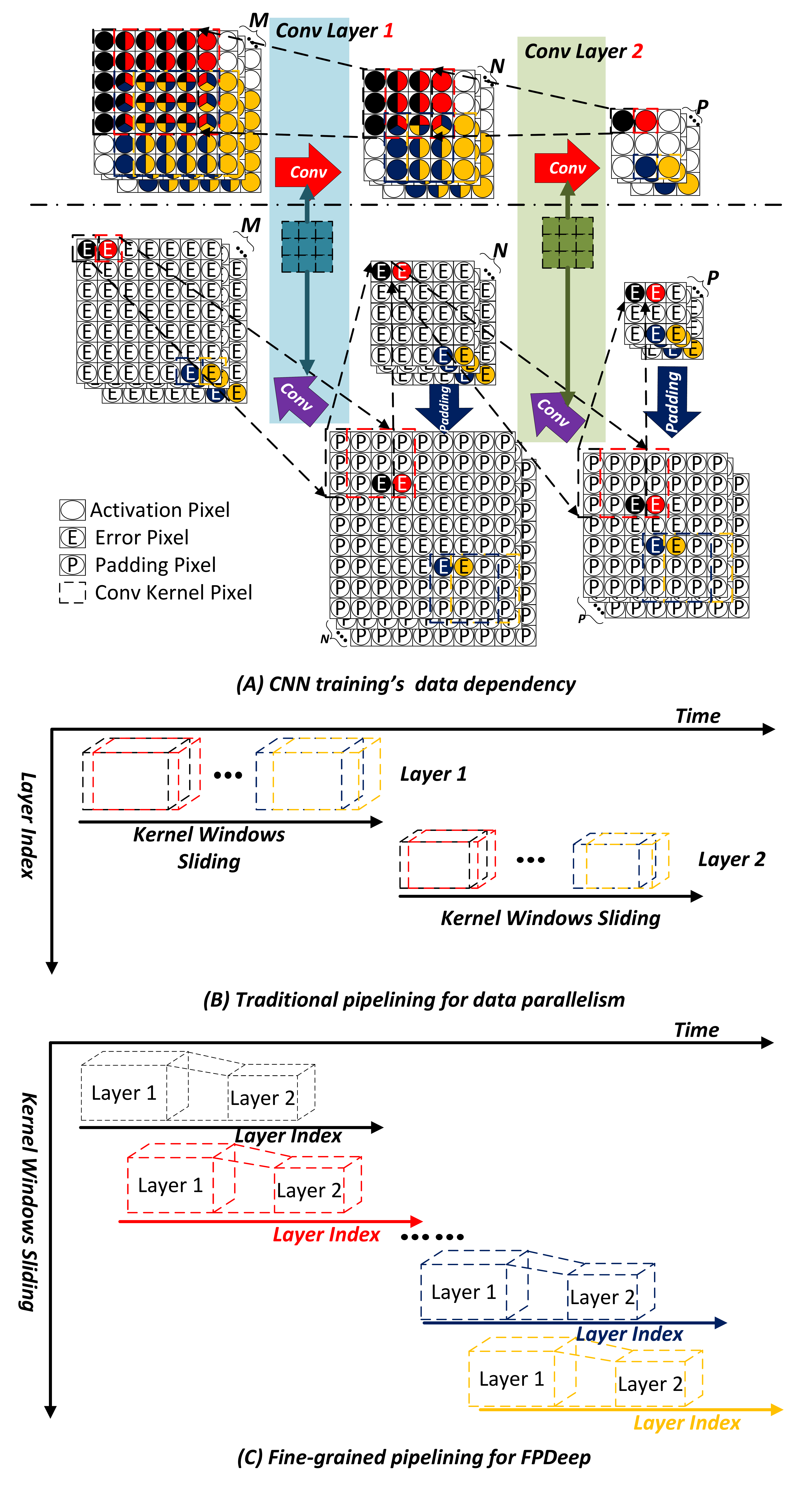}
 \vspace{-0.3truein}
\caption{FPDeep's fine-grained pipeline design showing (A) data dependencies of CNN training; (B) traditional data parallelism's coarse-grained pipeline; (C) FPDeep's fine-grained pipeline.}
\label{fig:depend}
\vspace*{0.05truein}
\end{figure}

To illustrate data dependencies during training we use as an example of two CONV layers with $3\times3$ kernel size. The operations of these two layers' forward/backward propagation are shown in Fig.\ref{fig:depend}(A). In forward propagation, a $7 \times 7$ feature map is fed into Layer 1 and a $5 \times 5$ feature map is generated. At layer 2 the $5 \times 5$ feature map is convolved with the parameters and inferred to a $3 \times 3$ feature map. In the backward propagation, the $3 \times 3$ error map is padded to $7 \times 7$ before it is fed to Layer 2. Next, the error map and corresponding parameters are convolved and another ($5 \times 5$) error map is produced; this is used for Layer 2's weight/bias gradient calculation. At Layer 1, the $5 \times 5$ error map is padded to $9 \times 9$ and then convolved to $7 \times 7$. 

The Fig.\ref{fig:depend}(A) depicts the data dependency of forwarding and backward propagations during CNN training. For the forward propagation phase, the image is inferred through all layers. To determine the data dependency, we start from the four activations at the output feature map at the last layer, which are marked as black, red, blue, and yellow, respectively, and trace backward to find the region of the input feature maps on which each depends. For the backward propagation phase, errors calculated are propagated backward through the network. To calculate gradients at a particular layer, errors which are backward propagated from the next layer and activations of its feature maps are necessary. Hence, the feature maps, which are generated in the forward propagation phase, need to remain available awaiting backward propagation. As shown in Fig.\ref{fig:depend}, activations, and errors among CONV layers show only fine-grained dependencies. That is, to begin computing the value of a pixel in a layer, only a fraction of the pixels from the previous layer are needed.  Therefore the computation of a layer can start much earlier before the previous layer is completely done. This provides the opportunity to process all CONV layers in parallel in a fine-grained pipeline.

The Fig.\ref{fig:depend}(B) shows the traditional method of accelerating CNN training. First, the $N$ channels of the feature maps are fed into the convolution layer $L1$.  Next, results from all $M$ channels begin to be processed while the convolution kernel slides across the $N$-channel feature maps. Much storage capacity is needed to maintain all temporal feature maps. Clearly, this method is not efficient. The fine-grained alternative is shown in Fig.\ref{fig:depend}(C): the calculation of an activation/error starts as soon as its dependent activations/errors are propagated from the previous/next layer. The basic process unit of FPDeep is an activation/error of a feature/error map; this is in contrast to the traditional method's basic unit of the entire feature/error map. The result is both a large increase in parallelism through the added pipeline stages and a reduction in storage so that only on-chip memory is needed.

\subsection{Parameter Alignment}

In contrast to traditional DP, where centralized gradient aggregation and weight update need to be performed sequentially, FPDeep conducts all of these processes in parallel. In order to achieve full hardware efficiency, we use a distributed and slightly-unaligned weight update scheme: gradient calculation, aggregation, and weight update are always performed locally. With respect to the parameter alignment, after the last training sample in a certain mini-batch (round $M$) is forward propagated through the cluster, its backward propagation follows immediately. At the same time, the forward propagation of training samples in the next mini-batch (round $M+1$) follows using the old parameters of round $M$ until the last training sample (from round $M$) is backward propagated. The deep fine-grained pipeline used in FPDeep guarantees fast feature and error propagation and reduces the time that old parameters need wait for the weight update, i.e. it eases the parameter alignment issue. Based on our experiments, with the mini-batch size as 1K and a cluster with 100 FPGAs, only the first 5 training samples of each epoch suffer from the resulting slight non-alignment. Moreover, this slight parameter non-alignment does not affect the convergence rate (as discussed in Section 6.4 and shown in Fig.\ref{fig:converge}(B)(D)(F)).

Existing work considers parameter alignment and high throughput to require a trade-off. Google’s GPipe \cite{huang2019gpipe} and Microsoft’s PipeDream \cite{narayanan2019pipedream} use a similar pipeline scheme to build a distributed DNN training system but use different alignment methods. GPipe divides the input mini-batch into several smaller micro-batches, enabling different GPUs/TPUs to work on different micro-batches simultaneously. GPipe needs to flush the pipeline and synchronize the gradients among all accelerators after the computation of the whole mini-batch finishes. GPipe’s solution introduces many bubbles in the pipeline. Moreover, GPipe focuses on fitting oversized DNN onto multiple accelerators, not solving the large-batch training problem. To the best of our knowledge, the approach used in GPipe makes the large-batch problem even more severe: more accelerators require more micro-batches and, in order to fill up each device, the size of micro-batches must be relatively large. As with GPipe, Microsofts’s PipeDream also uses coarse-grained workload partitioning and pipelining. However, in contrast to GPipe, PipeDream suffers from the parameter alignment issue. The authors propose a technique called {\it weight stashing} to save multiple versions of the parameters and so align parameters on a slightly longer time scale.

The optimization target of this paper is throughput, i.e. $epoch/h$. Note that in FPDeep, higher throughput is equivalent to reduced training execution time as FPDeep, because of the small mini-batch size, does not require more epochs to converge.

\section{Hardware Accelerator Architecture}
\begin{figure*}
\centering
\includegraphics[width=7.2in]{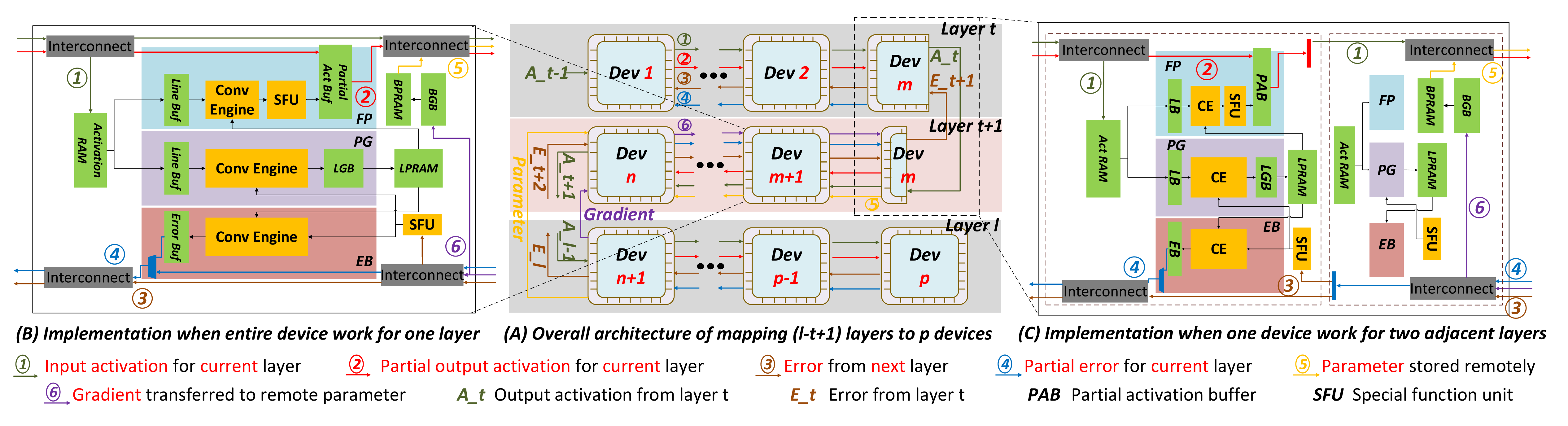}
\vspace*{-0.3truein}
\caption{Overall architecture of FPDeep accelerator and block design of each FPGA illustrating (A) the overall architecture of FPDeep; FPGAs can work cooperatively  on the same layer; also, multiple layers can be mapped on the same FPGA; (B) architecture of FPGA $m+1$, which is fully allocated to layer $t$; (C) architecture of FPGA $m$, which is allocated to both layer$t$ and layer $t+1$. }
\label{Arch}
\vspace*{0.05truein}
\end{figure*}


\subsection{Overall Architecture}

The overall architecture of the multi-FPGA accelerator and the detailed architecture of each FPGA are shown in Fig.~\ref{Arch}(B). For an $l$-layer CNN, FPDeep maps convolution layers $t \sim l$ to $p$ FPGAs. All $p$ FPGAs are connected in a 1-D topology.
In Fig.~\ref{Arch}, $A_t$ denotes the output activation of layer $t$ and $E_t$ the errors backward propagated from layer $t$. There are six key data-paths. Steps 1, 2 and 5 are for FP, while 3, 4 and 6 are for BP.

\noindent
{\bf 1.}  Output activations from layer $(t-1)$ are allocated to FPGAs of layer $t$ according to the ICP results. Each FPGA caches a segment of the features allocated to it and propagates the rest to the next node. 

\noindent
{\bf 2.}  Using the segment of features cached at Datapath 1, each FPGA calculates partial results of all output features at layer $t$. The partial features produced from FPGA $m$ are propagated to node $m+1$ through Datapath 2. After adding up partial features produced by nodes $m$ and $m+1$, the updated partial features are propagated to the next node.

\noindent
{\bf 3.}  In each cycle, errors from layer $(t+1)$ are back-propagated to the FPGAs of layer $t$ through Datapath 3.

\noindent
{\bf 4.}  Using errors from Datapath 3, each FPGA calculates the errors of the features allocated to it at Datapath 1 and propagates them to the preceding node. Node $m$ propagates the errors calculated by itself first and then the errors transferred from node $m+1$.

\noindent
{\bf 5.}  Parameters are transferred from the node where they are cached for parameter load balancing to the node where they are used to compute the output features. 

\noindent
{\bf 6.}  The gradients of parameters are transferred from the node where they are produced to the node where they are cached for parameter load balancing.

The proposed architecture is generally useful for SGD-based training of any feed-forward CNNs and can be extended to support other CNNs with more complex topologies such as ResNet and Inception. As described in Section 3.2, as long as a DNN can be described as a one-way graph with nodes representing pipelined groups of consecutive layers, FPDeep can efficiently partition and map its training logic to an FPGA cluster. New modules are needed as follows: {\it aggregation} to the filter concatenation in Inception; {\it  duplication of FP, PG, and EB} for the parallel CONV and Pooling kernels in Inception; and {\it gather and bypass} for the various types of shortcuts of ResNet. Integrating these into FPDeep is straightforward and will be part of the next-generation system.

\subsection{Single-FPGA Architecture}

As shown in Fig.\ref{Arch}, each FPGA includes FP, PG, and EB modules, as well as a memory subsystem to cache parameters, gradients, and activations. Each accelerator has 6 interconnection modules to communicate with its neighbors (this number is selected because it is available on many boards used for FPGA clusters and is sufficient for good scaling). An FPGA can be allocated to multiple layers. Implementations with FPGAs working for single layer and for multiple layers are illustrated in Fig.\ref{Arch}(B) and (C).

\subsubsection{Interconnection} 

There are two pairs of interconnection modules. 
{\bf a)} The upper pair, used by Forward datapaths 1, 2 and 5, 1) receives input features and partial features propagated by the preceding node; 2) bypasses the input features which are not mapped to it to the succeeding node; 3) adds partial results produced by FP to the received partial features and propagates updated partial features to the succeeding node; 4) forwards the parameters and gradients from the node which caches them to the node which produces them. 
{\bf b)} The bottom pair, used by backward datapaths 3, 4 and 6, 1) receives errors from the next layer bypassed by the succeeding node and passes them on to the preceding node; 2) receives errors of this layer calculated by the the succeeding node; 3) after errors are calculated by the EB module, propagates them to preceding node; 4) forwards the parameters and gradients from the node which calculates them to the node which caches them.

\subsubsection{Memory Subsystem}

The memory subsystem includes BRAM-based modules and stores activations, parameters, and gradients. 

\noindent
{\bf 1.} Activation RAM (Act-RAM) caches input activations mapped to the target FPGA until they are consumed in back-propagation and provides input activations as operators to FP and PG modules for output activation and parameter gradient calculation. After input activations are consumed in FP, they are kept in Act-RAM and wait to be reused during BP to calculate parameter gradients. Act-RAM is implemented as a FIFO-based memory. In BP, when errors are calculated and propagated backward from the adjacent FPGAs to a certain device, the features stored earlier in Act-RAM are first consumed by the PG module for parameter gradient calculations.

\noindent
{\bf 2.} Local Parameter RAM (LPRAM) caches parameters used as operators to produce the output activation at the local FPGA. For each FPGA, there are SF $\times$ K $\times$ K $\times$ OC parameters stored in LPRAM, where SF is the number of activations in the activation segment allocated to an FPGA. To provide enough concurrency of parameter access, LPRAM is designed as a SF $\times$ K $\times$ K-bank line buffer. Each bank caches OC parameters.

\noindent
{\bf 3.} Balanced Parameter RAM (BPRAM) caches the parameters mapped to the local device for parameter load balancing. These parameters are used as operators in other FPGA devices where on-chip memory is insufficient to cache all the required parameters. Both LPRAM and BPRAM are updated by PG. Similar to LPRAM, BPRAM is implemented as multi-bank line buffer. The number of banks and their depths are decided by the parameter balancing scheme. 

\noindent
{\bf 4.} Local Gradient Buffer (LGB) caches the gradients of the parameters stored in LPRAM. The gradients are cached and averaged at each iteration. At the point that a mini-batch size number of input figures are completely trained, the averaged gradients are forwarded to LPRAM and update the parameters stored in LPRAM.

\noindent
{\bf 5.} Balanced Gradient Buffer (BGB) caches the gradients of the parameters stored in BPRAM. These gradients are generated by and transferred from the device where the corresponding parameters are consumed.

\subsection{Forward Propagation (FP)}

The Line Buffer (LB) reads input features from the Act-RAM and feeds them to the Convolution Engines (CEs). The CEs perform convolutions with parameters from the LPRAM and input activation from the LB. In the Special Function Unit (SFU), the output features are activated, normalized, and sampled based on network specifications. Afterwards, features are transferred to the Partial Activation Buffer (PAB) and added to the partial features produced by and propagated by the preceding node. Finally, the updated partial activations are propagated to the next device through the interconnection module.



We use ICP as an example to show how the FP module calculates partial activation results. Assuming a certain FPGA node has been allocated with $S\_IC$ channels, at each cycle, $K \times K \times S\_IC$ activation are accessed from line buffers and broadcast to the CEs. Each CE consists of $S\_IC$ convolution tiles. Each convolution tile has $K \times K$ multiply-accumulate units and executes a $K \times K$ convolution operation per cycle. With all convolution tiles working on different input channels, at each cycle, each CE can finish calculating the partial results of one output channel. In the FP module, there are multiple CEs calculating the activations of different channels in parallel. The number of CEs, $P$, is determined by the number of DSPs allocated to FP operations during the offline ICP mapping. When partial activations are calculated, they are forwarded to partial activation buffers where they are used to update the partial results produced from the previous nodes.  

\subsection{Error Back-Propagation (EB)}

The EB module consumes errors from the next layer 
and produces errors for the target layer. 


It takes two steps to calculate the error of each input activation: (1) errors of all output feature maps are convolved, respectively, with their parameter filters and (2) their convolution outputs are summed. In FPDeep, for an FPGA allocated with $S\_IC$ input channels, the EB module calculates $S\_IC$ errors in parallel. In EB, CEs are used to perform the convolutions of errors of output channels and their parameters. This is different from the FP module where the number of convolution tiles in each CE is $S\_IC$, rather, each EB module has $S\_IC$ CEs. Each CE has $P$ tiles and each tile can perform a $K \times K$ convolution operation. The number of convolution tiles in each PE is pre-determined during ICP mapping. Taking $P$ as the number of CEs, at each cycle, the errors from $P$ from $OC$ output channels are broadcast to, and consumed by, $S\_IC$ CEs. The outputs are partial results of errors at $S\_IC$ input channels. After $OC/P$ cycles, all output channels are evaluated and the complete results of errors are forwarded to the Error Buffer. 


\subsection{Parameter Gradient Calculation (PG)}

PG consumes errors of the next layer propagated from the succeeding neighbor and calculates gradients of the parameters. Errors of output activations are used as convolution filters on the input activations which are cached in the Act-RAM during forward propagation. Gradients are cached in the Gradient Buffer and used to update parameters in LPRAM when a mini batch of samples is trained. 

In contrast to the convolutions at FP and EB where the filter size is normally smaller than 7, the filter sizes in PG ($R$ and $C$) can be in the hundreds. This requires expensive convolution tiles -- the resources can even exceed those of the FPGA. Even if this does not happen, the PG may still occupy most of the computing resources and result in a serious workload imbalance.
In PG, $K \times K$ $W \times W$ convolutions need to be performed. In FPDeep, we cut $K \times K$ large convolutions into $W \times W$ small convolutions ($K \times K$). The overall operation count stays the same. But in this case, the PG module always fits the DSP resources constraint and the CE array design of PG is similar to the ones in FP and EB.  


\section{Experiments and Evaluation}

In this section, we describe experiments performed to evaluate the efficiency of FPDeep. First, we evaluate the correctness and performance of the design on a small FPGA cluster with eight Xilinx VC709 boards. Based on these results, we validate a cycle-accurate software simulator. Because the small FPGA cluster is insufficient for complex neural networks, we use the software simulator to evaluate the performance of FPDeep on large-scale clusters.

\subsection{Small Scale Cluster Experiments}

\begin{figure}
\vspace*{-0.1truein}
   \centering
    \includegraphics[width=3.5in]{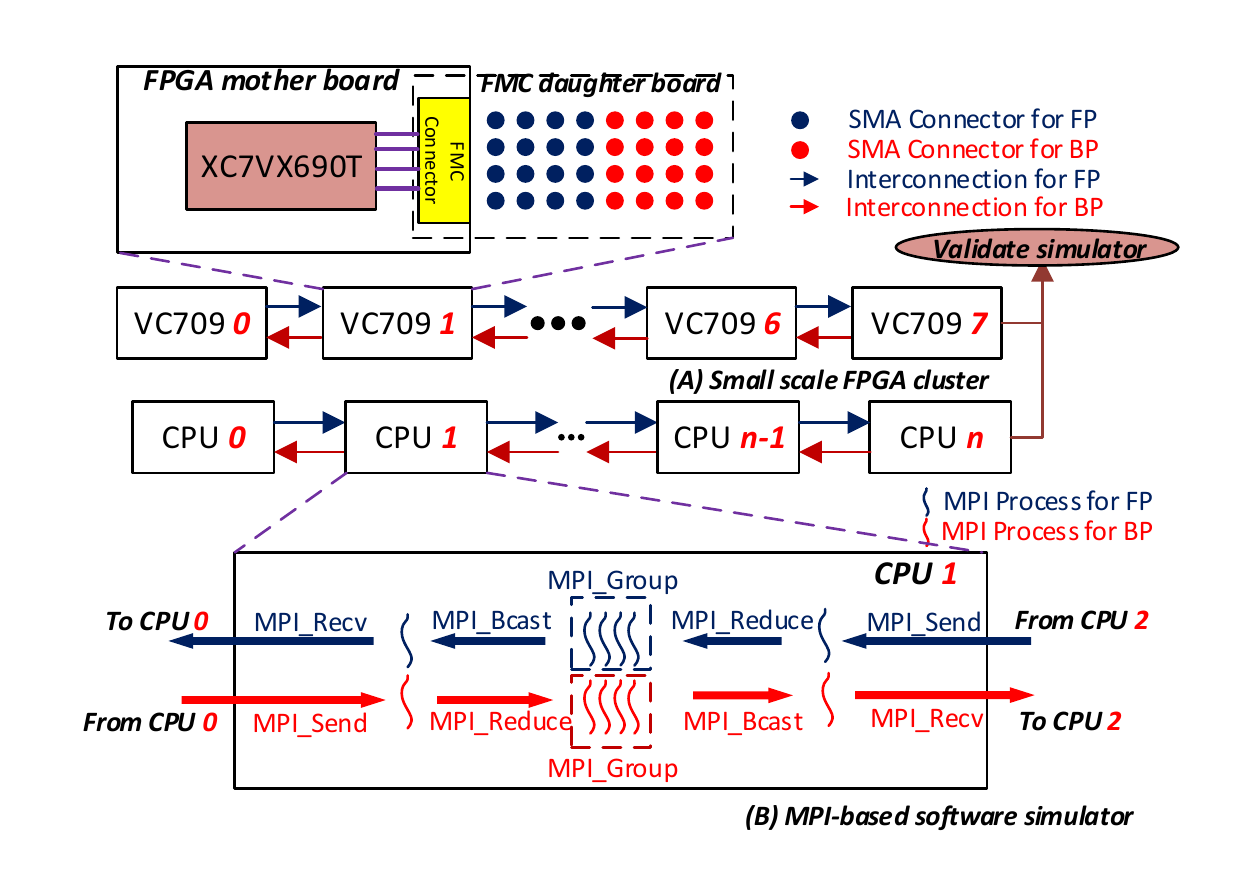}
\vspace*{-0.25truein}
    \caption{Hardware evaluation and MPI-based software simulator}
    \label{fig:exphs}
\vspace*{0.05truein}
\end{figure}

The small scale cluster experiments use a cluster of eight Xilinx VC709 evaluation boards. As shown in Fig. \ref{fig:exphs}(A), each VC709 motherboard contains one XC7VX690T FPGA, an FMC-HPC connector for the daughterboard extension, and an SMA FMC, which contains 32 SMA connectors. Some of the SMA connectors are used for forward propagation, others for backward propagation. The eight FPGA boards are connected in a 1-D daisy chain. This cluster is used to validate the parameterized hardware accelerators (Section 5), to perform the topology experiments (Section 4), and to validate the correctness of the software simulator.

\subsection{Large Scale Cluster Experiments}

The FPDeep software simulator is based on MPI version OpenMPI 2.1.1 (Fig. \ref{fig:exphs}); $n$ CPUs work in a pipelined manner. For each CPU, we have two MPI process groups, one each for FP and BP. Additional MPI processes handle data exchange with adjacent CPUs, broadcast the previous CPU's temporal activations, and reduce the current CPU's results. The simulator is parameterized to support various FPGA platforms. For example, the MPI processes which handle communication have configurable parameters that enable accurate simulation of the data exchange among FPGA boards with different interconnect bandwidths and latency.

The software simulator is currently used to evaluate large-scale VGG-19, VGG-16, and AlexNet training. For larger and more complex DNNs such as ResNet, GoogLeNet, and Recurrent Neural Networks, we will use emerging large-scale FPGA clusters. For example, the Open Cloud Testbed (associated with the Massachusetts Open Cloud), which was kicked off at the end of 2019, will in the first stage be equipped with at least 64 Xilinx Alveo 280 boards and be publicly available.

\subsection{Utilization and Performance}

\subsubsection{FPGA Resource Utilization}

\begin{figure*} 
\centering
\includegraphics[width=7.2in]{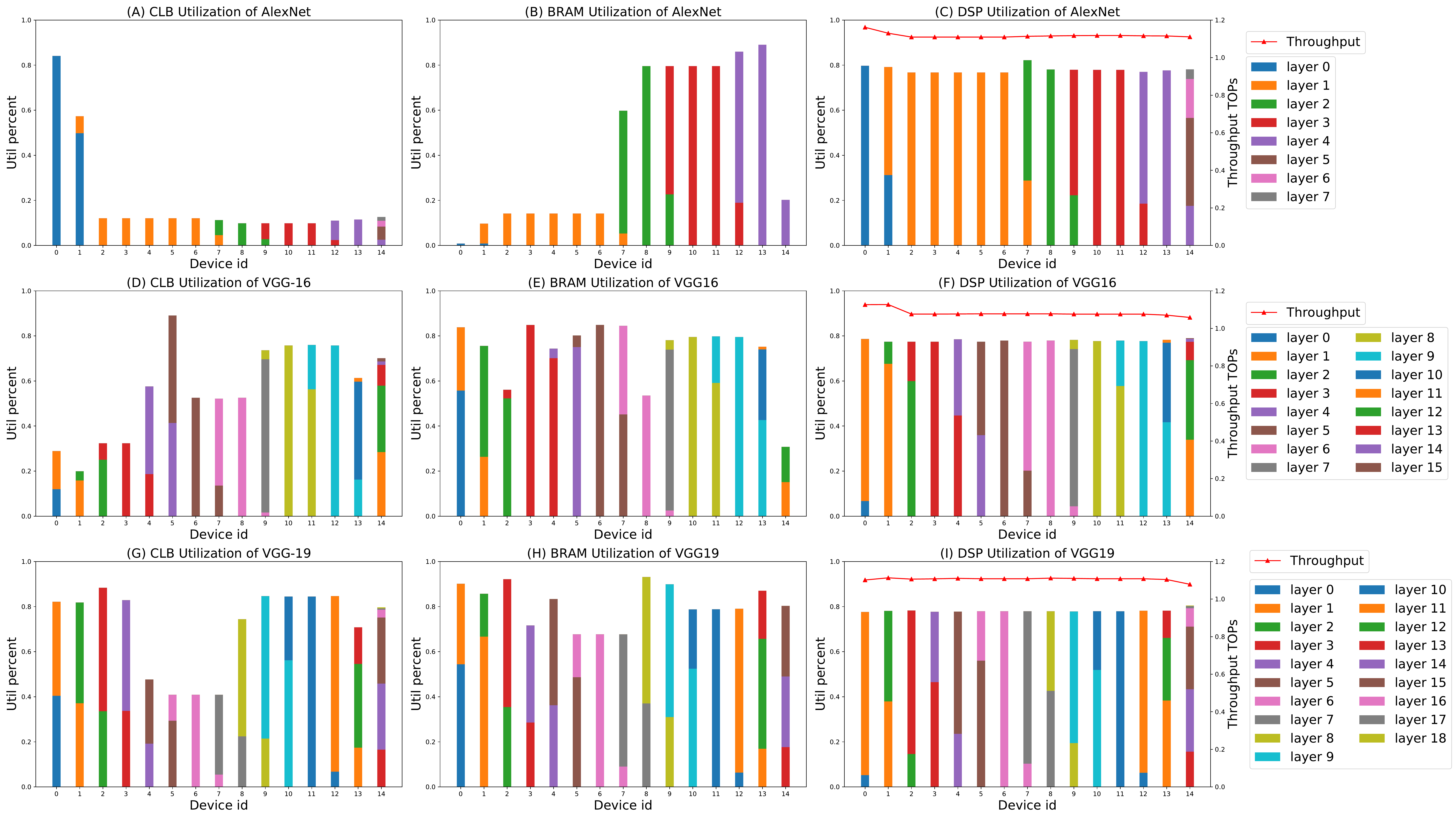}
\vspace*{-0.25truein}
\caption{Experimental results and utilization report when mapping AlexNet and VGG-16/19 to a cluster with 15 FPGAs}
\label{Result1}
\vspace*{0.05truein}
\end{figure*}

For illustration, we map AlexNet and VGG-16/19 to a cluster with 15 FPGAs. Figs. \ref{Result1}(A-I) show resource utilization of each FPGA and resource allocations among the network layers. As shown in the DSP utilization reports (Figs. \ref{Result1}(C)(F)(I)), the mapping is well-balanced. The utilization of DSP slices is roughly 80\% and the throughput of each FPGA is around 1 TOPS. On-chip BRAM is only used in the FPGAs that work solely on the CONV layers (FPGAs 1-14) and utilization of BRAMs is under 80\%. The highest bandwidth requirement among these 15 FPGAs for these three networks is 18.6 Gb/s.

For AlexNet (only 8 layers), the 15-node cluster does not require parameter balancing to achieve the best performance so CLB and BRAM utilization have been left unbalanced. For VGG-16/19, however, parameter balancing is required.

\subsubsection{Performance and Power Efficiency}

Table \ref{tab:exp} compares performance and power efficiency among the Titan X GPU \cite{zhang2016energy}, the Tesla K80 GPU \cite{tensoralex}, a previous FPGA implementation \cite{zhang2016energy}, and this work.

\cite{zhang2016energy} uses a workstation with an 8-core 3.8GHz AMD A10-5800K CPU and an Nvidia Titan X GPU. We use a server with Nvidia Tesla K80 GPUs as the golden model and baseline design. OpenBLAS and cuDNN libraries are used in software implementations.
\cite{zhang2016energy}'s CPU \& GPU and our GPU implementation are all based on data parallelism, while \cite{zhang2016energy}'s FPGA design is based on layer parallelism; the latter results in inter-board workload imbalance. The power consumption of all baseline and FPDeep systems are board-level and measured with a power meter.

FPDeep provides performance $5\times$ higher than previous FPGA work and comparable to the Titan X GPU. We evaluate energy efficiency with respect to GOPs/J. FPDeep provides $8.8\times$ better energy efficiency than the Titan X and $5.6\times$ better than the previous FPGA work. Compared with the K80, FPDeep provides $5.7\times$ better energy efficiency.

\begin{table*}
\caption{Cluster-level experimental results. All (CPU/GPU/FPGA) implementations use single precision floating point. \cite{zhang2016energy}, \cite{tensoralex}, and \cite{tensorvgg} do not give experiment results of training time per epoch} 
\vspace*{-0.1truein}
\centering
\begin{tabular}{|m{1.25cm}<{\centering}|m{1.3cm}<{\centering}|m{1.3cm}<{\centering}|m{1.3cm}<{\centering}|m{1.3cm}<{\centering}|m{1.5cm}<{\centering}|m{1.5cm}<{\centering}|m{1.5cm}<{\centering}|m{1.5cm}<{\centering}|m{1.5cm}<{\centering}|}
\hline
&CPU
\cite{zhang2016energy}&
GPU
\cite{zhang2016energy}&
\multicolumn{2}{|c|}{GPU}&
\multicolumn{2}{|c|}{FPGA\cite{zhang2016energy}}&
\multicolumn{3}{|c|}{FPDeep}\\
\hline
Device&
AMD A10&
Titan X	&
\multicolumn{2}{|c|}{Tesla K80}&
\multicolumn{5}{|c|}{Xilinx XC7VX690T}\\
\hline
CNN Model&	
AlexNet&
AlexNet&
AlexNet
\cite{tensoralex}&
VGG-16
\cite{tensorvgg}&
AlexNet&
VGG-16&
AlexNet&
VGG-16&
VGG-19	\\
\hline
Config&	
1 CPU&
1 GPU&
1 GPU&
1 GPU&
4 FPGAs&
1 FPGA&
15 FPGAs&15 FPGAs&15 FPGAs	\\
\hline
Perf (GOPS)&
34.23	&
1385	&
2330	&
2018    &
207	(Per FPGA)	&
290	(Per FPGA)	&
1157(Per FPGA)	&
1197(Per FPGA)	&
1220(Per FPGA)	\\
\hline
Training time (H)/ Epoch&
NA	&
NA	&
NA	&
NA    &
NA	&
NA	&
0.17	&
2.19	&
2.76	\\
\hline
Power efficiency (GOPS/J)	&
0.39	&
4.22	&
7.87	&
6.86    &
6.55	&
8.28	&
37.09	&
37.88	&
38.13	\\
\hline
\end{tabular}
\label{tab:exp}
\vspace*{0.05truein}
\end{table*}

\subsubsection{Load Balance and Optimization}

\begin{figure*}[ht!] 
\centering
\includegraphics[width=6.9in]{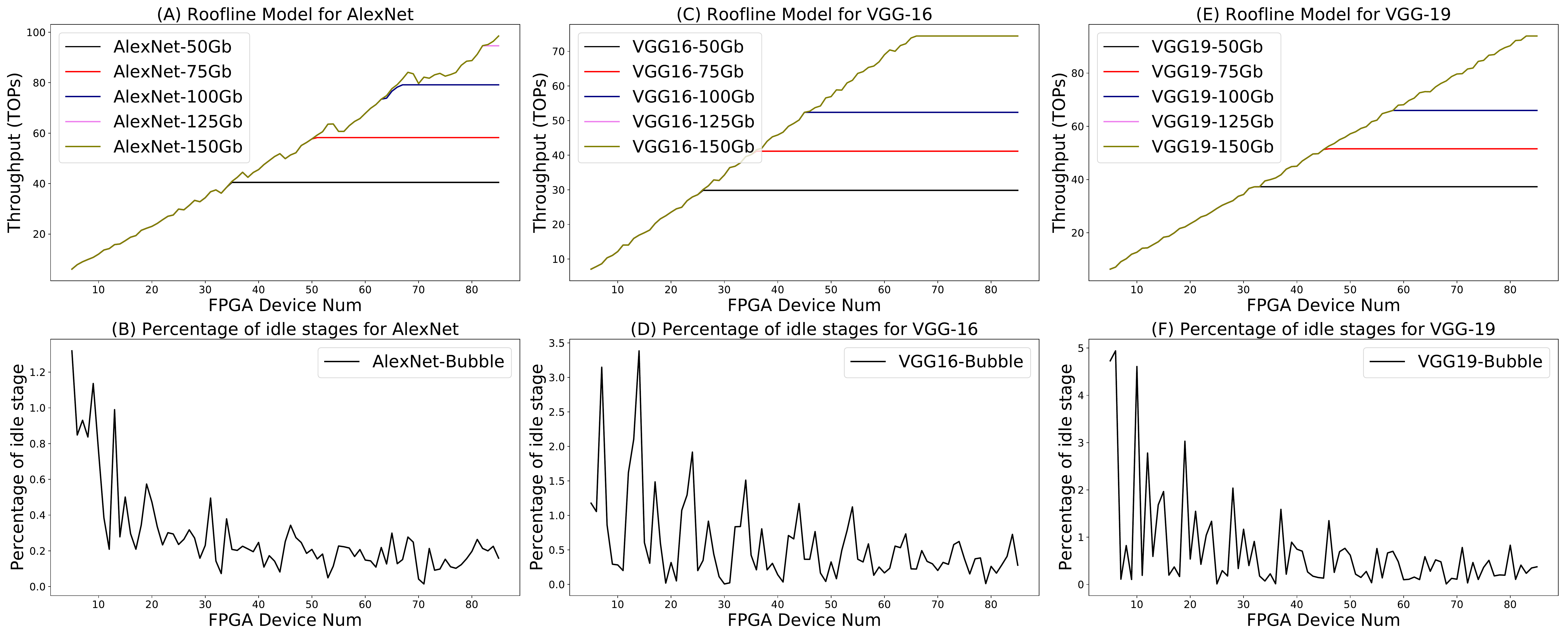}
\vspace*{-0.1truein}
\caption{Roofline models, percent idle stages, and epochs per hour of AlexNet, VGGNet-16, and VGGNet-19}
\label{Results}
\vspace*{0.05truein}
\end{figure*}

Alexnet, VGG-16, and VGG-19 are mapped onto clusters of sizes 5 to 85 with the cycle-accurate simulator. To demonstrate workload balance among FPGAs in different sized clusters, we present the proportions of idle stages. Figs. \ref{Results}(B)(D)(F) show that this is always under 5\%. When the number of FPGAs is more than 30, this number is stable with fluctuation between 0.5\% to 1\%. Generally, as the number of FPGAs increases, the proportion of idle stages decreases. The reason is that during ICP and OFP, the number of DSPs allocated to each layer is rounded to a multiple of $K\times K$. With more FPGAs and DSP resources, this effect is reduced. 

\begin{figure*}
    \centering
    \includegraphics[width=6.9in]{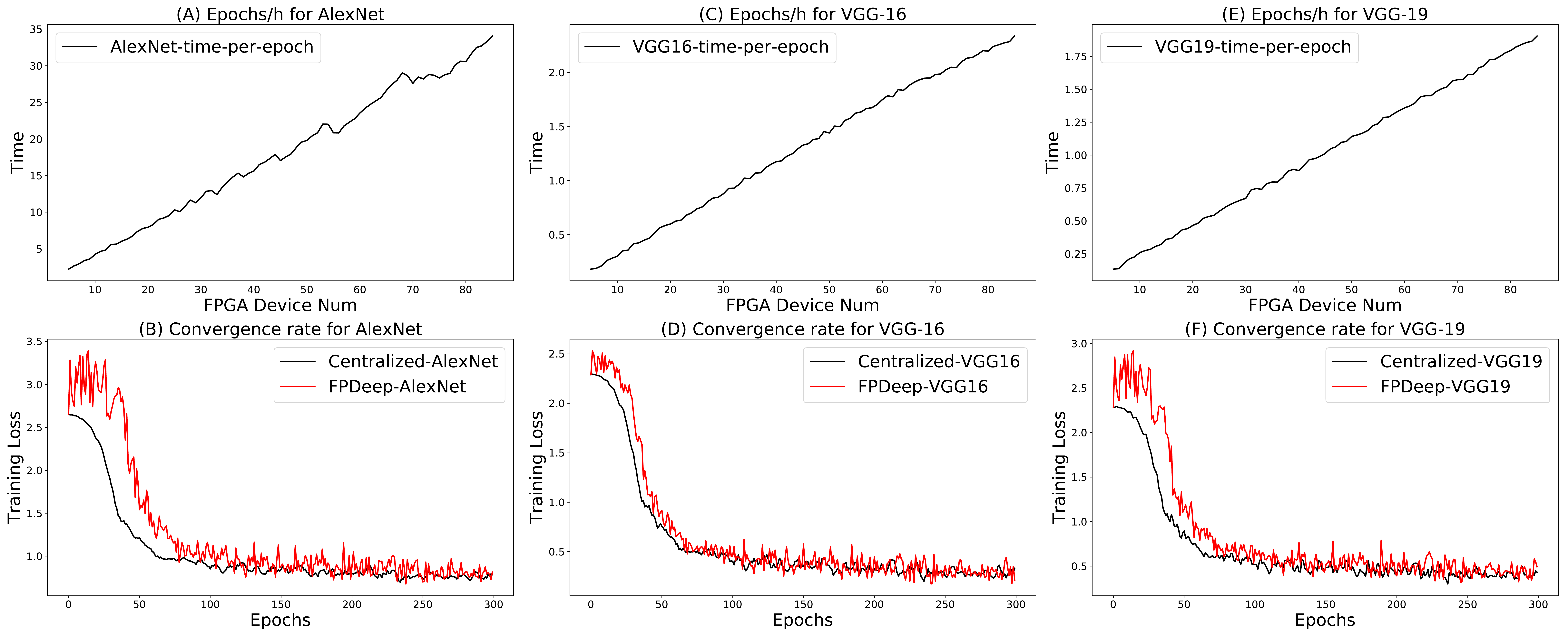}
\vspace*{-0.1truein}
    \caption{FPDeep's performance scalability and convergence rate.}
    \label{fig:converge}
 \vspace*{0.05truein}
 \end{figure*}

Computation and communication are critical constraints in system throughput. The roofline plots of AlexNet, VGG-16, and VGG-19 are shown in Figs. \ref{Results}(A)(C)(E). Note that the throughput has linear scaling up to the constraint imposed by inter-FPGA communication. For example, with 150 Gbps as the inter-board communication constraint, FPDeep shows linearity up to 83, 56, and 70 FPGAs for Alexnet, VGG-16 and VGG-19, respectively. As each transceiver (of that generation) can reach a maximum rate of 28 Gb/s, using 6 transceivers per FPGA achieves this number \cite{geng2019lp,sheng2018high,george2016novo,yang2019fully}.

Since high-end FPGAs frequently have more than 50 transceivers, scaling to much larger clusters is possible. The reason that bandwidth required by VGG-16 is larger than VGG-19 is straightforward: VGG-19 has more layers and thus more workload. During partitioning, with the same overall hardware resources, each layer of VGG-19 is allocated fewer resources. Thus, fewer batch features in each layer can be computed and transferred in parallel, which results in a smaller bandwidth requirement. 


\subsection{DNN Model Convergence}

Figs. \ref{fig:converge}(A)(C)(E) show the number of epochs that can be trained per hour. FPDeep provides a linear speedup of training per epoch. As hybrid model/layer parallelism does not constrain the choice of mini-batch size, the optimal learning rate and mini-batch size can always be applied in SGD, leading to the minimum number of epochs needed for training of a given accuracy. Hence, the linear speedup of training per epoch results in a linear speedup of CNN training.

Figs. \ref{fig:converge}(B)(D)(F) show the convergence rates of FPDeep and the traditional centralized DP training using the same small mini-batch size in SGD. The results show that FPDeep has similar convergence rates compared with the traditional centralized DP method, demonstrating that the slightly-unaligned weight update of FPDeep does not introduce additional training epochs. For the centralized case, we use a Sugon W740-G20 GPU server, which contains two Tesla K80 GPUs. The experiment is based on the Darknet framework; the CUDA library is cuDNN 5.0. For the FPDeep case, we use a Sugon CX50-G20 CPU cluster with an Intel Xeon E5-2680 v3 CPU. The FPDeep software simulator is compiled with gcc 7.1 and OpenMPI 2.1.1. The training dataset is CIFAR-10.


\section{Discussion and Future Work}
We propose a framework, FPDeep, which maps training logic of DNNs to multi-FPGA clusters with high efficiency and also automatically generates RTL implementations for target networks and clusters.

With FPDeep, clusters of FPGAs work in a deeply-pipelined manner using a 1-D topology; this enables the accelerators to map directly onto existing platforms, including Catapult, Catapult2, and almost any tightly-coupled FPGA cloud or cluster. FPDeep uses two mechanisms to facilitate high-performance and energy-efficiency: 1) various fine-grained partition and mapping strategies to balance workloads among FPGAs and 2) training of CNNs is executed in a fine-grained inter- and intra-layer pipelined manner, which reduces the time that features need for backward propagation and leads to a reduction in storage demand to the point where only on-chip memory is required for CONV layers. Experiments show that FPDeep has good scalability to a large number of FPGAs. The bottleneck is inter-FPGA communication bandwidth. However, we find that with 250 Gb/s bidirectional bandwidth per FPGA, which is easily supported by current generation FPGAs, FPDeep’s performance shows linearity up to 100 FPGAs. For example, using Alexnet and the VGGNets as benchmarks, with 6 transceivers per FPGA (e.g., using a 2014-era Altera Stratix-V), FPDeep shows linearity up to 83 FPGAs. We evaluate energy efficiency with respect to GOPs/J and find that FPDeep provides 5.7x to 8.8x higher energy efficiency than GPU servers.  

We briefly discuss future work. One area is supporting more complex NN models. Here, two additions are needed. First, while the current graph partitioning method supports ResNet and Inception (as described in Section 3.2), RNN and other new models require more complex graph structures. Second, support needs to be added for additional modules as described in Section 5.1. A second area is investigating benefits of hierarchical communication networks as arise when the nodes are multi-FPGA boards. Finally, another interesting question is use of off-chip memory. Currently, we only use off-chip memory when we are processing the fully connected layer. In the case of small clusters and large networks, where off-chip memory would be an option, we instead use the weight balancing scheme described in Section 3.4. In the future, as HBM becomes widespread, for very large networks it could make sense to use off-chip memory as an intermediate buffer to store activations and parameters.






\ifCLASSOPTIONcompsoc
  \section*{Acknowledgments}
\else
  \section*{Acknowledgment}
\fi

This research was partially supported by the DMC-CFA project and DS-HPC project under PNNL's Laboratory Directed Research and Development Program. It was also partially supported by the U.S. DOE Office of Science, Office of Advanced Scientific Computing Research, under award 66150: "CENATE - Center for Advanced Architecture Evaluation". The Pacific Northwest National Laboratory is operated by Battelle for the U.S. Department of Energy under contract DE-AC05-76RL01830. This work was also supported, in part, by the NSF through Awards CNS-1405695 and CCF-1618303/7960; by the NIH through Award 1R41GM128533; by grants from Microsoft and Red Hat; and by Intel through donated FPGAs, tools, and IP.

Preliminary versions of parts of this work appeared in FCCM 2018 \cite{geng2018fpdeep} and FPL 2018 \cite{geng2018framework}.

\ifCLASSOPTIONcaptionsoff
  \newpage
\fi

\bibliography{ref.bib}{}
\bibliographystyle{plain}

%
\vspace*{-0.35truein}
 
 \begin{IEEEbiography}[{\includegraphics[width=1in,height=1.25in,clip,keepaspectratio]{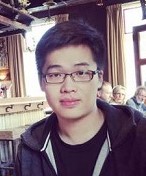}}]{Tong Geng}
received the BS degree from Zhejiang University, China in 2013 and the MS from Eindhoven University of Technology in 2015. He is working towards the PhD degree in Electrical \& Computer Engineering at Boston University. His research interests include machine learning, computer architecture, parallel computing, and reconfigurable computing.
 \end{IEEEbiography}
 
\vspace*{-0.35truein}

\begin{IEEEbiography}[{\includegraphics[width=1in,height=1.25in,clip,keepaspectratio]{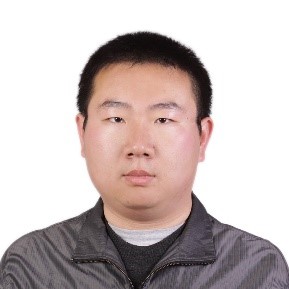}}]{Tianqi Wang}
received his Bachelor of Applied Physics from the University of Science and Technology of China in 2012 and is pursuing a PhD degree from University of Science and Technology of China. During 2018-2019 he was a visiting scholar at Boston University. Other interests include accelerating $N$-body problems on FPGA-centric clusters and using FPGAs in signal processing.
 \end{IEEEbiography}
 
\vspace*{-0.35truein}

\begin{IEEEbiography}[{\includegraphics[width=1in,height=1.25in,clip,keepaspectratio]{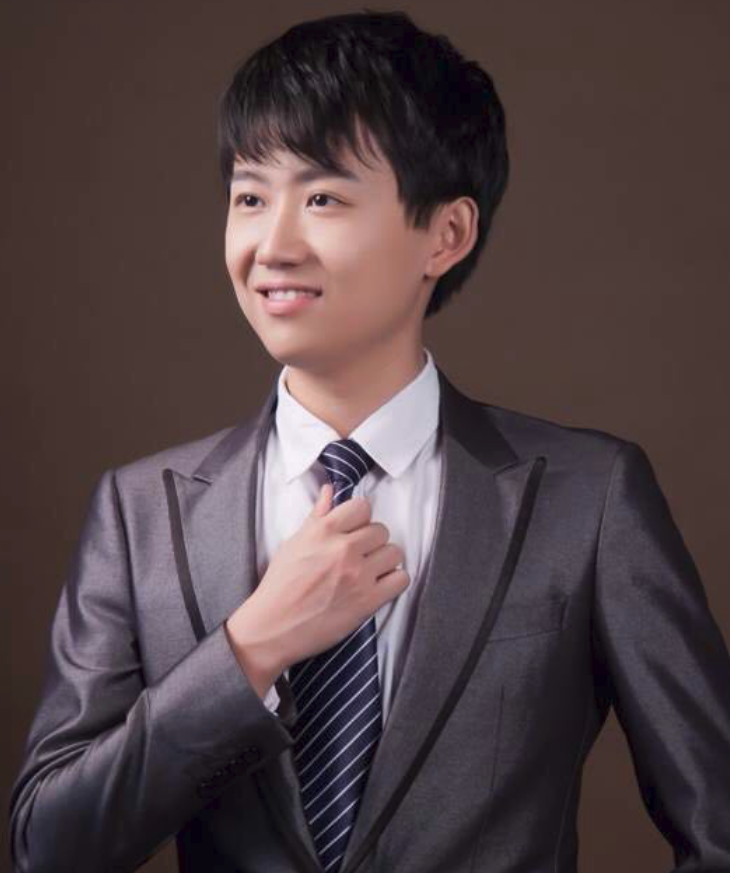}}]{Ang Li}
received his Bachelor degree from the CS department of Zhejiang University, Hangzhou, China, in 2010. From 2010 to 2012, he worked in industry as a software developer. In 2016, he received joint PhD degrees from the ECE department of National University of Singapore and the EE department of Eindhoven University of Technology. He is now a research scientist in the HPC group of the Pacific Northwest National Laboratory, USA.
\end{IEEEbiography}

\vspace*{-0.35truein}

\begin{IEEEbiography}[{\includegraphics[width=1in,height=1.25in,clip,keepaspectratio]{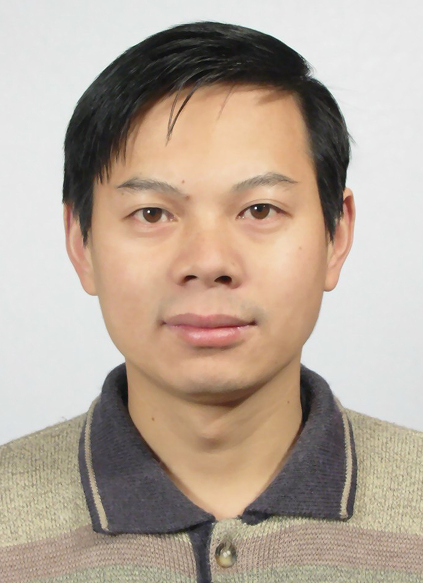}}]{Xi Jin}
is Associate Professor of Physics Department at University of Science and Technology of China where he directs the SoC Lab. His group has worked in accelerating scientific computing applications with FPGAs and using FPGA to build scientific instruments.
 \end{IEEEbiography}

\vspace*{-0.35truein}

\begin{IEEEbiography}[{\includegraphics[width=1in,height=1.25in,clip,keepaspectratio]{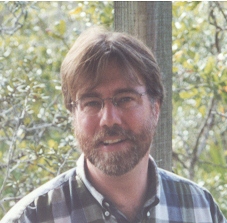}}]{Martin Herbordt}
is Professor of Electrical and Computer Engineering at Boston University where he directs the Computer Architecture and Automated Design Lab. His group has worked for many years in accelerating HPC applications with FPGAs and on system aspects of FPGA clusters and clouds.
\end{IEEEbiography}





\end{document}